\documentclass{article} % For LaTeX2e
\usepackage{iclr2026_conference,times}

\usepackage{amsmath,amsfonts,bm}

\def\eqref#1{equation~\ref{#1}}
\def\1{\bm{1}}

\DeclareMathAlphabet{\mathsfit}{\encodingdefault}{\sfdefault}{m}{sl}
\SetMathAlphabet{\mathsfit}{bold}{\encodingdefault}{\sfdefault}{bx}{n}

\usepackage[utf8]{inputenc} % allow utf-8 input
\usepackage[T1]{fontenc}    % use 8-bit T1 fonts
\usepackage{hyperref}       % hyperlinks
\usepackage{url}            % simple URL typesetting
\usepackage{booktabs}       % professional-quality tables
\usepackage{amsfonts}       % blackboard math symbols
\usepackage{nicefrac}       % compact symbols for 1/2, etc.
\usepackage{microtype}      % microtypography
\usepackage{xcolor}         % colors
\usepackage{natbib}         % citation package
\usepackage[font=small]{caption}
\usepackage{subcaption}

\usepackage{amsmath}
\usepackage{amssymb}
\usepackage{mathtools}
\usepackage{amsthm}

\usepackage{algorithm}
\usepackage{algorithmic}
\floatname{algorithm}{Algorithm}

\usepackage[capitalize,noabbrev]{cleveref}

\theoremstyle{plain}

\theoremstyle{definition}

\theoremstyle{remark}

\usepackage{tikz}
\usepackage{pgfplots}

\usepackage{multirow}
\usepackage{wrapfig}

\usepackage[table]{xcolor}
\definecolor{pastelImage}{HTML}{122571}        % light blue
\definecolor{pastelGex}{HTML}{F3A712}          % light green
\definecolor{pastelMulti}{HTML}{519872}        % light red/pink

\title{Learning Multimodal Representations from Partially Paired, Small-Scale Data}

\author{%
\textbf{Merel Kuijs}$^{*1,2}$ \quad
\textbf{Till Richter}$^{*1,2}$ \quad
\textbf{Rushin H Gindra}$^{*1,3,4,6}$ \\
\textbf{Korbinian Traeuble}$^{*1,2}$ \quad
\textbf{Christian Matek}$^{5}$ \quad
\textbf{Rebeka Lukn\'arov\'a}$^{1}$ \\
\textbf{Tingying Peng}$^{1,2,6}$ \quad
\textbf{Fabian J Theis}$^{1,2}$ \\[1em]
\begin{tabular}{@{}r@{\,}p{0.92\textwidth}@{}}
$^{1}$ & Institute of Computational Biology, German Research Center for Environmental Health, Helmholtz Zentrum M\"unchen, Neuherberg, Germany \\
$^{2}$ & Department of Computer Science, TUM School of Computation, Information and Technology, Technical University of Munich, Garching, Germany \\
$^{3}$ & Translational Cancer Research and Institute of Experimental Cancer Therapy, Klinikum rechts der Isar, TUM School of Medicine and Health, Technical University of Munich, Germany \\
$^{4}$ & Center for Translational Cancer Research, Technical University of Munich, Munich, Germany \\
$^{5}$ & Institute of Pathology, University Hospital Erlangen, Erlangen, Germany \\
$^{6}$ & Helmholtz AI, Helmholtz Zentrum M\"unchen, Neuherberg, Germany
\end{tabular} \\[1.5em]
Correspondence to: \texttt{fabian.theis@helmholtz-munich.de}
}

\iclrfinalcopy % Uncomment for camera-ready version, but NOT for submission.
\begin{document}

\maketitle

\renewcommand{\thefootnote}{\fnsymbol{footnote}}
\footnotetext[1]{Equal contribution.}
\renewcommand{\thefootnote}{\arabic{footnote}}

\begin{abstract}
Modeling multimodal data from partially paired samples is critical for advancing domains like biomedicine, where vast unimodal datasets and foundation models exist but paired data remains scarce. Existing fusion methods rely on large-scale paired datasets, limiting their use in scenarios with incomplete pairing. We introduce CAMEO, an adversarial learning-based modality fusion framework that integrates modalities from small, partially paired datasets. Combining any pre-trained unimodal encoder with a cross-modal latent alignment mechanism, CAMEO learns shared representations requiring only minimal paired samples. Evaluated on computational pathology tasks such as niche classification and cell type composition prediction, CAMEO achieves superior data efficiency, outperforming contrastive approaches like CLIP in low paired-data regimes, and highlighting the benefits of adversarial alignment when paired annotations are scarce. To facilitate further research, we additionally release a fully annotated HuggingFace dataset comprising three organs and paired image and gene expression modalities. By extending fusion methods to address limited pairing and small-scale datasets, we provide a framework that advances multimodal learning and broadens its applicability to real-world biomedical problems.

\end{abstract}

\section{Introduction}
\label{introduction}

Recent advances in spatial transcriptomics and histopathology (HP) imaging have advanced our understanding of tissue organization by providing complementary views of cellular states~\citep{larsson2021spatially, palla2022squidpy, fischer2023modeling}. While spatial transcriptomics captures gene expression with spatial information, HP imaging reveals tissue morphology and cellular organization at high resolution. The integration of these modalities promises deeper insights into tissue architecture and cellular interactions.

% b) Paragraph about unimodal foundation models and multimodal models on paired data
Foundation models have emerged as powerful tools for analyzing biomedical modalities~\citep{szalata2024transformers}. These models leverage self-supervised learning on large, unimodal datasets~\citep{richter2024delineating}. Models like scGPT~\citep{cui2024scgpt} and Geneformer~\citep{cui2024geneformer} learn gene-gene dependencies from gene expression data (i.e., quantitative transcriptomic profiles) through masked language modeling. This type of model is applicable to various cell- and gene-level downstream tasks, including cell annotation and gene function prediction. Vision models like UNI~\citep{uni} and CTransPath~\citep{ctrans} learn morphological patterns from H\&E-stained whole-slide images using a contrastive objective. % These objectives can either be distillation-based or discriminative, with each having their own advantages and limitations. 
However, these approaches do not utilize the potentially complementary information present in multiple modalities. While multimodal models have been successfully applied in domains such as language and vision~\citep{clip}, they typically require large-scale paired datasets, limiting their applicability to domains where paired data is scarce.

% c) Paragraph introducing our model
\begin{figure*}[t]
\vskip 0.2in
\begin{center}
\includegraphics[width=0.95\textwidth]{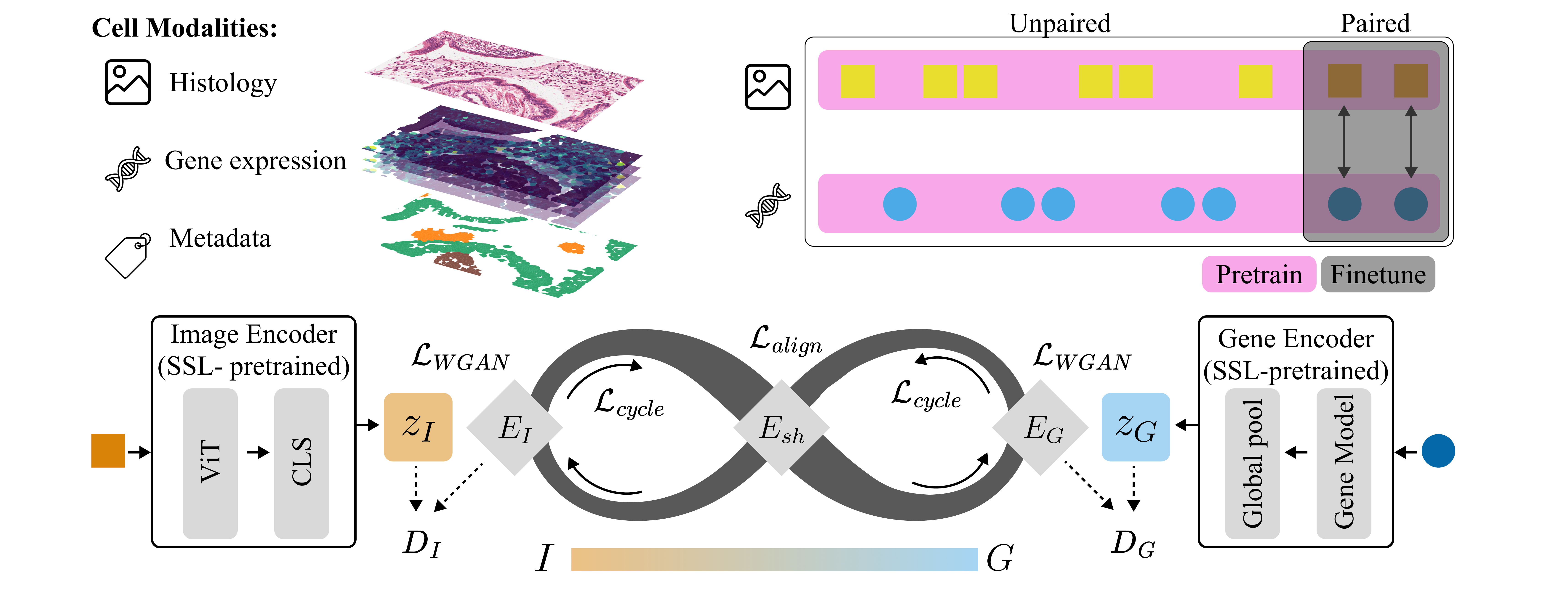}
\caption{CAMEO model. Top left: Cells in a tissue are represented by various modalities, including histology (vision), gene expression vectors (functional biology), and further metadata such as expert annotations that we use for downstream tasks. Top right: The model is first pre-trained on widely available, unpaired, unimodal data from vision (top) and gene expression (bottom) repositories, and then fine-tuned on small-scale paired datasets. Bottom: Unimodal image and gene expression encoders are trained with self-supervised learning to map images to $z_I$ and gene expression vectors to $z_G$. CAMEO then projects these generated representations into a shared multimodal latent space via $E_{sh}$. This allows prediction of the gene expression representation from the image representation and vice versa.}
\label{fig:model_concept}
\end{center}
\vskip -0.2in
\end{figure*}

% We introduce CAMEO (Fig.~\ref{fig:model_concept}), a multimodal model that combines self-supervised unimodal encoders with cross-modal alignment, enabling effective multimodal learning from small, partially paired datasets. This approach aligns latent spaces between gene expression and imaging data, enriching representations through artificial augmentations for downstream tasks.

% We introduce CAMEO (Fig.~\ref{fig:model_concept}), a GAN-based modality fusion framework that aligns representations from pre-trained unimodal encoders in a shared latent space. CAMEO's adversarial alignment mechanism effectively leverages small, partially paired datasets, making it particularly robust when limited paired samples are available compared to contrastive and other alignment approaches.

We introduce CAMEO (Fig.~\ref{fig:model_concept}), an adversarial modality fusion framework, as an alternative to contrastive and naive concatenation approaches. In biomedical settings where paired samples are scarce, CAMEO's adversarial mechanism provides robust performance with minimal paired data requirements. Our cross-tissue analysis reveals the complementary strengths of different fusion strategies, with adversarial alignment offering superior robustness when paired data is limited, while also highlighting the importance of systematically evaluating encoder combinations regardless of fusion approach.

% d) Paragraph listing the contributions
% We compare CAMEO to both unimodal baselines and multimodal approaches that rely on fully paired data, showing consistent improvements across downstream tasks. In summary, we present the following contributions:

Our comparative evaluation offers practical insights for choosing multimodal integration strategies in computational pathology. In summary, we present the following contributions:

\begin{itemize} 

%     \item We introduce CAMEO, a model that integrates unimodal self-supervised encoders with a cross-modal latent alignment mechanism for partially paired datasets.
    % \item We propose artificial augmentation through a shared latent space to enrich multimodal representations with limited paired data.
%     \item We release an aligned, expert-annotated HuggingFace dataset combining image and gene expression modalities.
%    \textcolor{pink}{\item We demonstrate that CAMEO outperforms unimodal baselines on two computational pathology tasks: niche annotation and cell-type composition prediction.
%     \item In scenarios characterized by significant distribution shifts, stemming from patient-based stratification and the resulting class heterogeneity, we show that CAMEO outperforms CLIP under low-pairing conditions.}
%     \item \textcolor{red}{We evaluate a range of unimodal baselines and multimodal encoder configurations, deriving practical recommendations for choosing the most effective encoder combinations when training multimodal models.}

\item We introduce CAMEO, an adversarial modality fusion framework that aligns pre-trained unimodal encoders in a shared latent space, effectively leveraging small, partially paired datasets.

\item We release an aligned, expert-annotated HuggingFace dataset combining image and gene expression modalities across three tissue types.

\item We demonstrate that multimodal approaches generally outperform unimodal baselines on niche classification tasks and that the ideal combination of encoders is dictated by the dataset.

\item We reveal that a unimodal encoder’s standalone accuracy is a poor predictor of its ultimate benefit in a multimodal setting. This highlights the need to systematically evaluate encoder combinations rather than assume that the best single-modality models will automatically yield the best fused models.

\item We show that CAMEO exhibits superior robustness to reduced pairing percentages compared to CLIP and concatenation baselines, particularly valuable in scenarios with limited paired data and class distribution shifts.

\end{itemize}

\section{Related Works}
\label{related}

\subsection{Multimodal Training on Paired Modalities}

Multimodal representation learning algorithms typically rely on paired data to align embeddings across modalities. For instance, CLIP \citep{clip} is trained on a massive corpus of 400 million image-text pairs, showcasing impressive performance in zero-shot classification tasks. Similarly, ImageBind \citep{imagebind} aligns multiple modalities under the condition that samples are represented by at least one common modality, which can be restrictive when such overlaps are scarce. BLIP-2 \citep{blip2} aligns visual features with a pre-trained language model through a Querying Transformer, leveraging paired image-text data for training. Similarly, LLaVa \citep{llava} fine-tunes pre-trained models on paired image-text data to enable complex reasoning. CoCa \citep{coca}, on the other hand, is trained from scratch, utilizing paired data to support multimodal tasks like captioning and cross-modal alignment. 

\subsection{Multimodal Approaches in Pathology}
In pathology, multimodal methods have leveraged the integration of imaging and textual data to improve performance in diagnostic and prognostic tasks. These approaches generally fall into two regimes: patch-level and slide-level. Patch-level vision-language methods \citep{plip, conch, pathchat, pathclip, quilt1m} typically adopt CLIP-style contrastive pretraining. In contrast, slide-level methods such as TITAN \citep{titan}, PathAlign \citep{pathalign}, and PRISM \citep{prism} leverage multiple instance learning (MIL) algorithms on paired whole-slide image and text data to perform, for example, pathology report generation and cross-modal retrieval of histology slides and clinical annotations.

In addition to image-text data, image-expression data has also been explored in multimodal pathology approaches. Tangle \citep{tangle} combines histology images with paired Bulk RNA sequencing data for few-shot slide classification and retrieval, showing improved performance over unimodal baselines on slide-level tasks. This highlights the benefit of integrating high-dimensional sequencing data with imaging data. RNA-CDM \citep{rna-cdm}, HE2RNA \citep{HE2RNA}, and BLEEP \citep{bleep} explore the translation between imaging and sequencing modalities, generating one modality from instances of the other. \citet{Yangetal} uniquely enable this translation with unpaired data. Survival prediction tasks have also benefited significantly from multimodal approaches, as exemplified by MOTCat \citep{motcat}, CMTA \citep{cmta}, GPDBN \citep{gpdbn}, and HFBSurv \citep{HFBSurv}. These methods integrate paired imaging and sequencing data to build predictive models that outperform unimodal counterparts in terms of clinical outcome prediction. More recent methods, such as MUSK \citep{musk}, relax the need for fully paired data by pre-training on large-scale unpaired datasets. While achieving strong performance in precision oncology tasks, MUSK’s dependence on large corpora for robust cross-modal embeddings may limit its generalizability when only a small amount of data is available.

\subsection{Modal Alignment and Integration Techniques}
% Within the biomedical domain, there has been interest in fusing radiology or histopathology images with genomic or transcriptomic data \citep{Mobadersany2018, tist}. % However, many efforts still rely on extensive patient-level pairing.  Methods like scVI \citep{scvi} and TotalVI \citep{totalvi} leverage deep probabilistic models to integrate single-cell omics data, but do not address alignment with imaging modalities. In contrast, methods like Pathomic Fusion \citep{pathofusion} and PathOmics \citep{pathomics} are specifically designed to integrate paired imaging and omics data. Pathomic Fusion applies the Kronecker product to fuse paired unimodal embeddings, while PathOmics aligns paired embeddings in a shared latent space using a mean squared error loss.
Recent alignment work has sought to integrate unpaired multimodal samples by incorporating prior knowledge or causal inference techniques. GLUE \citep{glue} employs a knowledge graph to guide the integration of unpaired multi-omics data, offering an interpretable framework for combining disparate modalities. However, its generalizability is constrained when constructing an appropriate knowledge graph is impractical or infeasible. \citet{xi2024} estimate a propensity score in each modality, capturing shared information between latent states and perturbations. Though this method enables matching unpaired samples across modalities, it still requires robust data curation. Another line of research has focused on using optimal transport to align unpaired omics datasets. Methods such as SCOT \citep{demetci2022scot} and GROT \citep{grot} leverage the shared gene feature space to achieve cross-modal alignment, but gene expression and histology images lack a common coordinate system, making it difficult to define a biologically meaningful transport cost between them. \citet{nakada} extend the CLIP loss to incorporate unpaired data, enabling alignment of paired images sampled from MNIST and Fashion-MNIST. This synthetic dataset might oversimplify multimodal relationships, limiting the generalizability of the findings to more complex natural data distributions.

Our approach builds on this growing body of multimodal alignment research. However, we focus on real-world applications in clinical pathology where data is limited and paired samples are scarce due to the expense of obtaining them. We specifically target patch-level tasks rather than slide-level analyses. Since the latter rely on MIL algorithms, they are unsuitable for patch-level tasks. Unlike previous patch-level methods trained on paired vision-language datasets, our approach leverages both paired and unpaired vision-transcriptomics data. By using pre-trained unimodal encoders and a cross-modal alignment mechanism, we enable efficient adaptation to downstream tasks with minimal paired labeled examples. This capability is particularly relevant in medical settings, where collecting large amounts of paired multimodal data can be both costly and time-consuming. Beyond demonstrating effectiveness on downstream tasks, we conduct systematic ablation studies across a range of gene expression and image encoder combinations. We find that multimodal models typically achieve better performance than unimodal baselines, though results depend on the combination of encoders used. Importantly, the effectiveness of unimodal encoders alone does not consistently forecast their success within multimodal models, emphasizing the need for thorough evaluation when choosing encoders for multimodal approaches.
\section{Method}

\subsection{Overview}

% Removed to save space
% CAMEO is built on the hypothesis that histology images and gene expression profiles provide distinct but complementary information -- morphological context and functional biology, respectively. Integrating these modalities is typically challenging, since paired samples (matching histology images and gene expression profiles from the same tissue region) are scarce and expensive to obtain, while unpaired data is relatively abundant but affected by batch effects and technical variations. Here, we present a general adversarial framework for learning aligned representations between partially paired modalities through adversarial training. Our results demonstrate that this framework effectively leverages the limited paired data to learn representations that are both biologically meaningful and predictive, outperforming unimodal baselines. While our primary application is to histology and gene expression data, this framework is broadly applicable to other multimodal datasets.

CAMEO is modular by design and compatible with any image or gene expression encoder. It follows a three-stage process: (1) initial data embedding with modality-specific encoders, (2) unsupervised representation alignment via cross-modal generation, and (3) supervised evaluation via linear probing for downstream tasks. By using pre-computed latent embeddings instead of raw data, we reduce computational complexity and memory requirements. CAMEO's design allows us to incorporate powerful pretrained encoders, such as foundation models trained on large-scale histology or single-cell data, which can improve robustness and generalizability. Our approach addresses several key challenges in multimodal self-supervised learning, including limited data availability, which makes assembling large datasets costly and technically difficult; scarcity of paired data, which constrains models that require fully paired examples such as CLIP; and batch effects, arising from systematic variations introduced by different experimental protocols and equipment.
% \begin{itemize}
%     \setlength{\itemsep}{2pt}  
%     \setlength{\parskip}{1pt}
%     \setlength{\parsep}{1pt}
%     \item \textbf{Data Scarcity}: Collecting large datasets can be challenging due to cost and technical constraints,
%     \item \textbf{Paired Data Availability}: Paired datasets are scarce, limiting the ability to train models that require fully paired data, such as CLIP,
%     \item \textbf{Batch Effects}: Different experimental protocols and equipment introduce systematic variations.
% \end{itemize}

Next, we provide a detailed description of the individual components of the model: 1) the vision encoder, 2) the gene expression encoder, 3) the pretraining process using multimodal fusion, and (4) the linear probing stage for classification and regression.

\subsection{Vision Encoders}
\label{background_img}
Let $x^{I} \in \mathbb{R}^{H \times W \times 3}$ denote an image patch extracted from a tissue slide, where $H$ and $W$ are the height and width of the patch, and the 3 channels correspond to RGB color intensities. Each image shows the morphological structure of the cells present at an average magnification of $20$x and a resolution of $0.5$ microns per pixel~(MPP). 
Modern foundation models for pathology often adopt Vision Transformer (ViT)~\citep{vit16x16} architectures pre-trained on large cohorts of histology slides. To evaluate the role of the vision encoder in multimodal fusion, we conduct an ablation using five pretrained models (four ViT-based encoders and one convolutional encoder). These encoders are used in a plug-and-play fashion within the CAMEO framework. Further details on the specific encoders used are provided in Section \ref{exp_baseline}.

% \textcolor{red}{We adopt UNI~(ViT-Large or ViT-L)~\citep{uni} as our foundational image encoder. UNI was trained on patches from a collection of $\sim100K$ tissue slides, termed the `Mass-100K' dataset, as well as the Genotype-Tissue Expression~(GTEx)~\citep{gtex} dataset. The model is pretrained using the DINOv2~\citep{dinov2} framework, which utilizes a knowledge distillation based self-supervision objective. Without requiring further fine-tuning, UNI has demonstrated the ability to extract robust image features for diverse downstream tasks, making it suitable for our application.}
% \textcolor{red}{We test multiple pretrained image encoders to test performance consistency CAMEO}

\subsection{Gene Expression Encoders}
\label{background_gex}

Let ${X^G} \in \mathbb{R}^{n \times p}$ denote the gene expression matrix, where $n$ is the number of cells or spots and $p$ is the number of measured genes. Each input $x^{G} \in \mathbb{R}^p$ is a high-dimensional vector representing the expression profile of a single cell or spot. % Spatial transcriptomics incorporates spatial coordinates, capturing gene expression and positional cell or spot neighborhood relationships. 
Gene expression encoders transform these vectors into compact latent representations that capture biological signals relevant for downstream tasks.

% For the Lung and Breast datasets as shown in \ref{exp_data}, we model the cell expressions in a graph where the data is represented as a graph $G=(V,E)$. Each node (cell) $v \in V$ has an associated gene expression profile $x_v \in \mathbb{R}^p$. Edges $E$ encode spatial proximity given a radius $r$~\citep{fischer2023modeling}. For the Thymus data set as shown in \ref{exp_data}, we use the spot gene expression matrix as is, to input to the models.

To systematically assess how different gene encoding strategies impact performance, we conduct an ablation over five gene expression encoders. This includes four existing models from the literature, as well as a Graph Attention Network (GAT) that we train ourselves~\citep{richter2023spatialssl}. The GAT encoder follows the formulation of Graph Attention Networks~\citep{velickovic2017graph}, learning node embeddings by aggregating information from spatial neighbors using learned attention weights. We pretrain the GAT using a self-supervised masked modeling objective: random subsets of each expression vector are masked and reconstructed from unmasked entries using the neighborhood structure. All encoders are used in a plug-and-play manner, and details on their usage and evaluation setup are provided in Section~\ref{exp_baseline}.

\subsection{Multimodal Fusion}
In this work, we propose a general adversarial framework for learning cross-modal representations.

\subsubsection{Model Architecture}
The adversarial fusion module consists of two cross-modal generators ($\mathcal{G}$) and their corresponding discriminators ($\mathcal{D}$):
\begin{align*}
\mathcal{G}_{G\rightarrow I}: \mathcal{X}_G &\rightarrow \mathcal{X}_I & \mathcal{D}_I: \mathcal{X}_I &\rightarrow \mathbb{R} \\
\mathcal{G}_{I\rightarrow G}: \mathcal{X}_I &\rightarrow \mathcal{X}_G & \mathcal{D}_G: \mathcal{X}_G &\rightarrow \mathbb{R}
\end{align*}
Here, $\mathcal{X}_I \subset \mathbb{R}^{d_I}$ and $\mathcal{X}_G \subset \mathbb{R}^{d_G}$ denote the embedding spaces of images and gene expression, respectively. The inputs to the generators are $z_I \in \mathbb{R}^{d_I}$ and $z_G \in \mathbb{R}^{d_G}$, which are obtained from modality-specific unimodal encoders:
\[
z_I = F_I(x^I), \quad z_G = F_G(x^G),
\]
where $F_I: x^I \rightarrow \mathbb{R}^{d_I}$ is the image encoder and 
$F_G: x^G \rightarrow \mathbb{R}^{d_G}$ is the gene expression encoder.
Each generator employs a shared bottleneck architecture:
\begin{align*}
\mathcal{G}_{G\rightarrow I}(z_{G}) &= E_{sh}(E_G(z_{G})) \\
\mathcal{G}_{I\rightarrow G}(z_{I}) &= E_{sh}(E_I(z_{I}))
\end{align*}
where $E_G$ and $E_I$ are projection layers that adapt expression and image embeddings, respectively, before passing them into the shared layers $E_{sh}$. % In this way, an embedding from one modality can be translated into the latent space of the other modality, allowing us to derive a corresponding image embedding from gene expression or an expression embedding from an image.

\subsubsection{Training Objectives}
\label{method_training}
\paragraph{Stage 1: Representation Learning}
In the first stage, we focus on learning shared representations from partially paired image and gene expression data. Given paired samples $\{(x_i^I, x_i^G)\}_{i=1}^{n_p}$ and unpaired samples $\{x_j^I\}_{j=1}^{n_u^I}$, $\{x_k^G\}_{k=1}^{n_u^G}$, our goal is to learn mappings between modalities that preserve semantic relationships.

% CAMEO's objective combines three components: (i) an adversarial loss, where discriminators encourage generators to produce realistic cross-modal embeddings, (ii) a cycle-consistency loss, which enforces semantic preservation during modality translation, and (iii) an alignment loss, which explicitly matches paired examples in a shared latent space. The total loss is a weighted sum of these terms; detailed formulations are provided in Appendix~\ref{app:objectives}.

The objective of CAMEO consists of three components:

1. \textbf{Adversarial Loss}: For each modality, a discriminator learns to distinguish real from generated embeddings~\citep{gan}:
   \begin{align*}
   \mathcal{L}_{WGAN} &= \mathbb{E}_{z_I \sim p_I}[\mathcal{D}_I(z_I)] - \mathbb{E}_{z_G \sim p_G}[\mathcal{D}_I(\mathcal{G}_{G\rightarrow I}(z_G))] \\
   &+ \mathbb{E}_{z_G \sim p_G}[\mathcal{D}_G(z_G)] - \mathbb{E}_{z_I \sim p_I}[\mathcal{D}_G(\mathcal{G}_{I\rightarrow G}(z_I))]
   \end{align*}
% Here, $z_I = F_I(x^I)$ and $z_G = F_G(x^G)$ are embeddings obtained from modality-specific encoders $F_I$ and $F_G$. $\mathcal{G}_{I\rightarrow G}$ and $\mathcal{G}_{G\rightarrow I}$ are cross-modal generators, while $\mathcal{D}_I$ and $\mathcal{D}_G$ are the corresponding discriminators. The 1-Lipschitz constraints on $D_I$ and $D_G$ are enforced via weight clipping.

2. \textbf{Cycle Consistency Loss}: To ensure semantic preservation during cross-modal translation~\citep{cyclegan}:
   \begin{align*}
   \mathcal{L}_{cycle} &= \mathbb{E}_{z_I \sim p_I}[\|\mathcal{G}_{G\rightarrow I}(\mathcal{G}_{I\rightarrow G}(z_I)) - z_I\|_1] \\
   &+ \mathbb{E}_{z_G \sim p_G}[\|\mathcal{G}_{I\rightarrow G}(\mathcal{G}_{G\rightarrow I}(z_G)) - z_G\|_1]
   \end{align*}

3. \textbf{Alignment Loss}: For paired samples $(z_I^p, z_G^p)$ from the set of paired examples $\mathcal{P}$, embeddings are explicitly aligned in the shared latent space:
   \begin{equation*}
   \mathcal{L}_{align} = \mathbb{E}_{(z_I^p, z_G^p) \sim \mathcal{P}}[\|E_{sh}(z_I^p) - E_{sh}(z_G^p)\|_2^2]
   \end{equation*}

The total loss combines these terms with weighting factors:
\begin{equation*}
\mathcal{L}_{total} = \lambda_1\mathcal{L}_{cycle} + \lambda_2\mathcal{L}_{WGAN} + \lambda_3\mathcal{L}_{align}
\end{equation*}

Formally, Stage~1 optimization follows a min–max formulation:
\begin{equation*}
% \min_{E_I,E_G,E_{sh},G_{I\rightarrow G},G_{G\rightarrow I}} \; \max_{D_I,D_G} \; 
% \lambda_1 \mathcal{L}_{cycle} + \lambda_2 \mathcal{L}_{WGAN} + \lambda_3 \mathcal{L}_{align}
\min_{\mathcal{G}_{I\rightarrow G}, \mathcal{G}_{G\rightarrow I}} 
\big( \lambda_1 \mathcal{L}_{cycle} + \lambda_3 \mathcal{L}_{align} \big) + 
\min_{\mathcal{G}_{I\rightarrow G}, \mathcal{G}_{G\rightarrow I}} \; 
\max_{\mathcal{D}_I, \mathcal{D}_G} \lambda_2 \mathcal{L}_{WGAN}.
\end{equation*}

During Stage~1, the generators $\mathcal{G}_{I\rightarrow G}$ and $\mathcal{G}_{G\rightarrow I}$ are updated using $\mathcal{L}_{total}$, while the discriminators $\mathcal{D}_I$ and $\mathcal{D}_G$ are updated using only $\mathcal{L}_{WGAN}$.

\paragraph{Stage 2: Linear Probing}

In the second stage, we train a linear probing head on the paired data. We keep the generators $\mathcal{G}_{I\rightarrow G}$, $\mathcal{G}_{G\rightarrow I}$ frozen. These generators are used solely to produce embeddings for each gene expression and image sample in the shared latent space. The resulting embeddings from both modalities are then concatenated and passed to the linear probing head for classification or regression:
\begin{equation*}
\hat{y} = h_\theta([E_{sh}(E_I(z^I)); E_{sh}(E_G(z^G))])
\end{equation*}

Only the parameters $\theta$ of the linear probing head are updated during this stage, using a task-specific loss -- mean squared error (MSE) for regression or cross-entropy for classification.
% Removed to save space
% \begin{equation*}
% \mathcal{L}_{task} = \begin{cases}
%     -\sum_i y_i \log \hat{y}_i & \text{classification} \\
%     \|y - \hat{y}\|_2^2 & \text{regression}
% \end{cases}
% \end{equation*}
Definitions of the macro F1 score, micro F1 score, and $R^2$ metric used for evaluation are provided in Appendix~\ref{app_metrics}. CAMEO provides a stable, deterministic embedding that accommodates limited paired data and non-coinciding manifold structures (see Appendix~\ref{app_prob_align}).

\section{Experiments}
\label{experiments}

In this section, we evaluate each multimodal framework's ability to learn from partially paired, small-scale data. Our experiments address two key questions: (1) whether these representations outperform unimodal approaches that leverage only image or gene expression data, and (2) how CAMEO compares to multimodal baselines that require larger amounts of paired data. 
% We present our datasets (Section~\ref{exp_data}), computational pathology tasks (Section~\ref{exp_tasks}), baselines (Section~\ref{exp_baseline}), and results spanning three tissues (breast, lung, and thymus) (Section~\ref{exp_results}), focusing on performance in low-data pairing scenarios. 
We assess performance across a wide range of encoder combinations, varying both gene expression and vision backbones in a plug-and-play manner. In addition, we examine how performance changes under different pairing ratios, ranging from fully paired to extremely limited paired data, allowing us to isolate the impact of cross-modal alignment under realistic constraints. We describe the datasets in Section~\ref{exp_data}, the tasks in Section~\ref{exp_tasks}, the baselines in Section~\ref{exp_baseline}, and report quantitative results spanning three tissues (breast, lung, and thymus) in Section~\ref{exp_results}. Throughout, we emphasize the role of encoder choice, data availability, and task characteristics in shaping the success of multimodal learning strategies.

% In this section, we evaluate CAMEO's ability to learn meaningful multimodal representations from partially paired, small-scale data. Our experiments address two key questions: (1) whether these representations outperform unimodal approaches that leverage only image or gene expression data, and (2) how CAMEO compares to multimodal baselines that require larger amounts of paired data. We first introduce our datasets and experimental setup (Section~\ref{exp_data}), then describe two computational pathology tasks: niche classification and cell-type composition prediction (Section~\ref{exp_tasks}). We benchmark against various unimodal and multimodal baselines (Section~\ref{exp_baseline}), demonstrating CAMEO's effectiveness particularly in low-data pairing scenarios (Section~\ref{exp_results}). Our analysis spans two distinct tissue types, breast and lung, allowing us to assess the generalizability of our approach across different biological contexts.

\subsection{Data}
\label{exp_data}
For the evaluation of multimodal learning frameworks and unimodal baselines, we use two 10x Xenium and one 10x Visium spatial transcriptomics cohorts. Each dataset includes expert-generated ground-truth annotations, but predicting these labels is challenging due to intra-class heterogeneity, class imbalance, and the need to account for spatial and contextual dependencies. This makes them well-suited for methods that integrate complementary information from two independent modalities. The Lung Pulmonary Fibrosis (LungPF) dataset comprises 23 10x Xenium samples~\citep{lpf}, collected to investigate the pathobiology of PF.
% Removed to save space
% To our knowledge, LungPF is the largest imaging-based spatial transcriptomics study of the human lung, with detailed histological and cell type annotations.
A second 10x Xenium cohort includes 7 publicly available breast cancer samples from 4 patients, hosted on the 10x Genomics website and annotated by our collaborating pathologist. The third dataset is a 10x Visium cohort of 19 thymus samples~\citep{thymus} from 11 donors (6 pediatric, 5 fetal), used to study T cell development during fetal and early postnatal stages. To construct HuggingFace (HF) datasets for multimodal learning, we aligned gene expression and imaging modalities by linking each image patch to its corresponding gene expression profile. All datasets will be shared with the research community. We split each into training, validation, and test sets, stratified by patient (see Appendix~\ref{app_datasets}).

\subsection{Downstream Tasks}
\label{exp_tasks}
% Removed to save space
% Models were evaluated on two downstream tasks: a histology-based niche classification task and a transcriptomics-based cell type composition prediction task. The objective was to demonstrate that cross-modal training outperforms unimodal baselines.
\subsubsection{Niche classification}
Niches are biologically relevant tissue regions that regulate cell behavior, typically involving multiple cell types and extracellular components. Niche annotation involves identifying these regions based on expert-labeled ground truth, enabling detailed tissue characterization. This is crucial for understanding region-specific processes, such as immune infiltration or stromal activity, which are linked to prognosis and treatment outcomes. In this task, image patches are treated as niches and classified into $n_{classes}$ different classes based on tissue morphology, as annotated by a pathologist. For the LungPF cohort, $24$ histological niches were identified, each representing a distinct pulmonary fibrotic tissue region. Similarly, for the breast cohort, $14$ distinct histological features were annotated. For the thymus cohort, $11$ regions were annotated based on morphological landmarks \citep{thymus}. % \textcolor{red}{Further details and distributions of the annotations are shown in App Figure \ref{fig:data_dist} and Table \ref{tab:sample-summary}.}
Further details and distributions of the annotations are shown in Appendix~\ref{app_datasets}.
\subsubsection{Cell type composition prediction}
Many diseases, including cancer and autoimmune disorders, involve changes in cell type composition. Estimating cell type composition from gene expression data enables the identification of disease-associated diagnostic bio-markers and provides insight into the disease microenvironment, which is crucial for developing targeted therapies. In the cell type composition prediction task, the goal is to estimate the fraction of each cell type present in a patch. We evaluate our model on predicting the composition of $9$ broad cell types defined by the group that created the LungPF cohort. % \textcolor{red}{Further details are available in App Figure \ref{fig:data_dist} and Table \ref{tab:sample-summary}.} 
Further details are available in Appendix~\ref{app_datasets}.

\begingroup
\setlength{\tabcolsep}{3pt}
\renewcommand{\arraystretch}{0.9}

\begin{table*}[ht]
\scriptsize
    \centering
    \caption{Multi‐class niche classification performance comparison of models on the Lung ($n_{classes} = 24$), Breast ($n_{classes} = 14$), and Thymus ($n_{classes} = 11$) datasets using \textbf{100\% paired data}. An asterisk (*) denotes that the model was not evaluated on the test set. The GAT model is only compatible with single-cell spatial data. As a result, we exclude the GAT encoder from experiments on the Thymus dataset, which contains spot-based gene expression measurements. Blue, orange, and green text indicate image encoders, gene expression encoders, and multimodal methods, respectively.}
    \label{tab:model_comparison}
    \resizebox{\textwidth}{!}{%
    \begin{tabular}{@{}llcccccc@{}}
        \multirow{2}{*}{Data Type} & \multirow{2}{*}{Model Name} 
            & \multicolumn{2}{c}{Lung} 
            & \multicolumn{2}{c}{Breast} 
            & \multicolumn{2}{c}{Thymus} \\ 
        \cmidrule(lr){3-4} \cmidrule(lr){5-6} \cmidrule(lr){7-8}
        & & Macro\ F1  & Micro\ F1 
          & Macro\ F1 & Micro\ F1 
          & Macro\ F1 & Micro\ F1\\
        \midrule
        \multirow{5}{*}{Image} 
        & \textcolor{pastelImage}{CONCH}      & 0.099 & 0.286 & 0.169 & 0.197 & 0.152 & 0.211 \\
        & \textcolor{pastelImage}{CTransPath} & 0.105 & 0.304 & 0.128 & 0.188 & 0.127 & 0.218 \\
        & \textcolor{pastelImage}{ResNet}     & 0.096 & 0.246 & 0.104 & 0.156 & 0.136 & 0.205 \\
        & \textcolor{pastelImage}{UNI}        & 0.134 & \underline{0.401} & 0.133 & 0.133 & 0.127 & 0.194 \\
        & \textcolor{pastelImage}{Virchow}    & 0.116 & 0.391 & 0.145 & 0.146 & 0.158 & 0.222 \\
        \midrule
        \multirow{5}{*}{Gene Expression} 
        & \textcolor{pastelGex}{CellPLM}     & 0.183 & 0.199 & 0.130 & 0.369 & 0.134 & 0.227 \\
        & \textcolor{pastelGex}{GAT}         & 0.171 & 0.263 & 0.181 & 0.383 & *     & *     \\
        & \textcolor{pastelGex}{Nicheformer} & 0.158 & 0.211 & 0.134 & 0.242 & 0.102 & 0.183 \\
        & \textcolor{pastelGex}{scGPT}       & 0.177 & 0.269 & 0.153 & 0.386 & 0.134 & 0.208 \\
        & \textcolor{pastelGex}{scVI}        & 0.125 & 0.234 & 0.144 & 0.312 & 0.164 & 0.253 \\
        \midrule
        \multirow{27}{*}{Multimodal}
        % CAMEO group
        & \textcolor{pastelMulti}{CAMEO}-\textcolor{pastelImage}{CTransPath}-\textcolor{pastelGex}{GAT}         & 0.166 & 0.177 & 0.096 & 0.335 & *     & *     \\
        & \textcolor{pastelMulti}{CAMEO}-\textcolor{pastelImage}{CTransPath}-\textcolor{pastelGex}{Nicheformer} & 0.075 & 0.064 & 0.080 & 0.233 & 0.193 & \textbf{0.347} \\
        & \textcolor{pastelMulti}{CAMEO}-\textcolor{pastelImage}{CTransPath}-\textcolor{pastelGex}{scVI}        & 0.158 & 0.223 & 0.079 & 0.294 & 0.155 & 0.214 \\
        & \textcolor{pastelMulti}{CAMEO}-\textcolor{pastelImage}{CONCH}-\textcolor{pastelGex}{GAT}              & 0.128 & 0.287 & 0.124 & 0.374 & *     & *     \\
        & \textcolor{pastelMulti}{CAMEO}-\textcolor{pastelImage}{CONCH}-\textcolor{pastelGex}{Nicheformer}      & 0.092 & 0.253 & 0.186 & 0.277 & 0.207 & 0.290 \\
        & \textcolor{pastelMulti}{CAMEO}-\textcolor{pastelImage}{CONCH}-\textcolor{pastelGex}{scVI}             & 0.164 & 0.278 & 0.154 & \textbf{0.445} & \underline{0.212} & 0.293 \\
        & \textcolor{pastelMulti}{CAMEO}-\textcolor{pastelImage}{UNI}-\textcolor{pastelGex}{GAT}                & 0.137 & 0.274 & 0.098 & 0.227 & *     & *     \\
        & \textcolor{pastelMulti}{CAMEO}-\textcolor{pastelImage}{UNI}-\textcolor{pastelGex}{Nicheformer}        & 0.193 & 0.308 & 0.141 & 0.297 & 0.172 & \underline{0.302} \\
        & \textcolor{pastelMulti}{CAMEO}-\textcolor{pastelImage}{UNI}-\textcolor{pastelGex}{scVI}               & 0.137 & 0.249 & 0.181 & 0.411 & 0.185 & 0.267 \\
        % CLIP group
        & \textcolor{pastelMulti}{CLIP}-\textcolor{pastelImage}{CTransPath}-\textcolor{pastelGex}{GAT}          & 0.199 & \textbf{0.412} & 0.204 & 0.407 & *     & *     \\
        & \textcolor{pastelMulti}{CLIP}-\textcolor{pastelImage}{CTransPath}-\textcolor{pastelGex}{Nicheformer}  & 0.188 & 0.330 & 0.225 & 0.382 & 0.176 & 0.255 \\
        & \textcolor{pastelMulti}{CLIP}-\textcolor{pastelImage}{CTransPath}-\textcolor{pastelGex}{scVI}         & 0.182 & 0.352 & 0.209 & 0.419 & 0.202 & 0.269 \\
        & \textcolor{pastelMulti}{CLIP}-\textcolor{pastelImage}{CONCH}-\textcolor{pastelGex}{GAT}               & 0.180 & 0.358 & 0.195 & 0.341 & *     & *     \\
        & \textcolor{pastelMulti}{CLIP}-\textcolor{pastelImage}{CONCH}-\textcolor{pastelGex}{Nicheformer}       & 0.158 & 0.309 & 0.196 & 0.341 & 0.173 & 0.247 \\
        & \textcolor{pastelMulti}{CLIP}-\textcolor{pastelImage}{CONCH}-\textcolor{pastelGex}{scVI}              & 0.157 & 0.315 & 0.184 & 0.342 & 0.161 & 0.248 \\
        & \textcolor{pastelMulti}{CLIP}-\textcolor{pastelImage}{UNI}-\textcolor{pastelGex}{GAT}                 & 0.195 & 0.382 & 0.194 & \underline{0.422} & *     & *     \\
        & \textcolor{pastelMulti}{CLIP}-\textcolor{pastelImage}{UNI}-\textcolor{pastelGex}{Nicheformer}         & 0.177 & 0.334 & 0.133 & 0.332 & 0.170 & 0.251 \\
        & \textcolor{pastelMulti}{CLIP}-\textcolor{pastelImage}{UNI}-\textcolor{pastelGex}{scVI}                & 0.194 & 0.335 & 0.170 & 0.305 & 0.188 & 0.259 \\
        % Concat group
        & \textcolor{pastelMulti}{Concat}-\textcolor{pastelImage}{CTransPath}-\textcolor{pastelGex}{GAT}        & \textbf{0.214} & 0.368 & \underline{0.293} & 0.337 & *     & *     \\
        & \textcolor{pastelMulti}{Concat}-\textcolor{pastelImage}{CTransPath}-\textcolor{pastelGex}{Nicheformer}& \underline{0.203} & 0.399 & 0.218 & 0.266 & 0.187 & 0.283 \\
        & \textcolor{pastelMulti}{Concat}-\textcolor{pastelImage}{CTransPath}-\textcolor{pastelGex}{scVI}       & 0.165 & 0.300 & 0.240 & 0.241 & \textbf{0.227} & 0.301 \\
        & \textcolor{pastelMulti}{Concat}-\textcolor{pastelImage}{CONCH}-\textcolor{pastelGex}{GAT}             & 0.162 & 0.350 & \textbf{0.305} & 0.359 & *     & *     \\
        & \textcolor{pastelMulti}{Concat}-\textcolor{pastelImage}{CONCH}-\textcolor{pastelGex}{Nicheformer}     & 0.181 & 0.298 & 0.140 & 0.187 & 0.176 & 0.276 \\
        & \textcolor{pastelMulti}{Concat}-\textcolor{pastelImage}{CONCH}-\textcolor{pastelGex}{scVI}            & 0.108 & 0.277 & 0.250 & 0.300 & 0.179 & 0.253 \\
        & \textcolor{pastelMulti}{Concat}-\textcolor{pastelImage}{UNI}-\textcolor{pastelGex}{GAT}               & 0.145 & 0.392 & 0.152 & 0.233 & *     & *     \\
        & \textcolor{pastelMulti}{Concat}-\textcolor{pastelImage}{UNI}-\textcolor{pastelGex}{Nicheformer}       & 0.130 & 0.398 & 0.140 & 0.141 & 0.148 & 0.198 \\
        & \textcolor{pastelMulti}{Concat}-\textcolor{pastelImage}{UNI}-\textcolor{pastelGex}{scVI}              & 0.126 & 0.322 & 0.136 & 0.199 & 0.152 & 0.190 \\
        \bottomrule
    \end{tabular}%
    }
\end{table*}
\endgroup

\subsection{Baselines}
\label{exp_baseline}
We evaluate three categories of baselines: image-only, gene expression-only, and multimodal. Image baselines use five pretrained encoders: UNI~\citep{uni}, CTransPath~\citep{ctrans}, ResNet-50~\citep{resnet}, Virchow ~\citep{virchow}, and CONCH~\citep{conch}. These models are used in a frozen state, and a linear classifier is trained on top (linear probing) to assess the quality of their representations. Gene expression baselines take high-dimensional gene expression vectors as input. We apply linear probing to CellPLM~\citep{cellplm}, scGPT~\citep{cui2024scgpt}, Nicheformer~\citep{nicheformer}, and the GAT encoder described in Section~\ref{background_gex}, keeping all encoders frozen. For scVI~\citep{scvi}, we train the encoder from scratch using only the training split of our paired HuggingFace datasets, and apply a linear classifier on top of the learned embeddings. 
% We consider four frameworks: CLIP~\citep{clip}, Concatenated, UIMC~\citep{uimc}, and MM-LEGO~\citep{mm-lego}.
For multimodal baselines, we consider two frameworks: CLIP~\citep{clip} and ``Concat''. In the ``Concat'' framework, the raw output embeddings from the unimodal gene expression and image encoders are concatenated directly, without any explicit alignment or fusion mechanism. We test each multimodal framework using different combinations of gene expression and image encoders. All multimodal frameworks produce joint embeddings that are used to train a task-specific linear head. Further details for all baselines are provided in Appendix~\ref{app_model_details}.

\begin{figure}[ht]
\vskip -0.2in
\begin{center}
\setlength{\tabcolsep}{1pt}  % reduce spacing between columns
\begin{tabular}{ccc}
\includegraphics[width=0.34\textwidth]{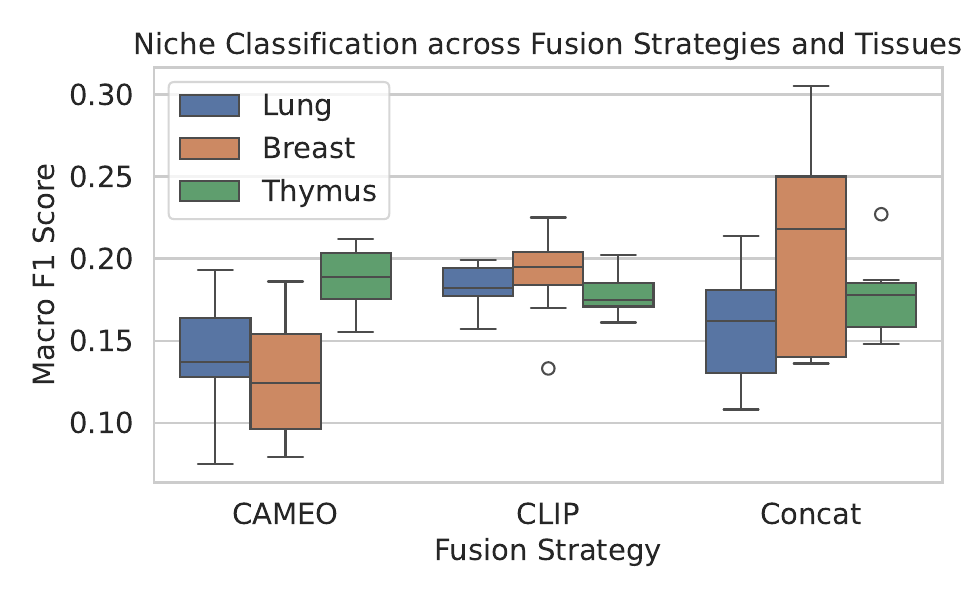} &
\includegraphics[width=0.34\textwidth]{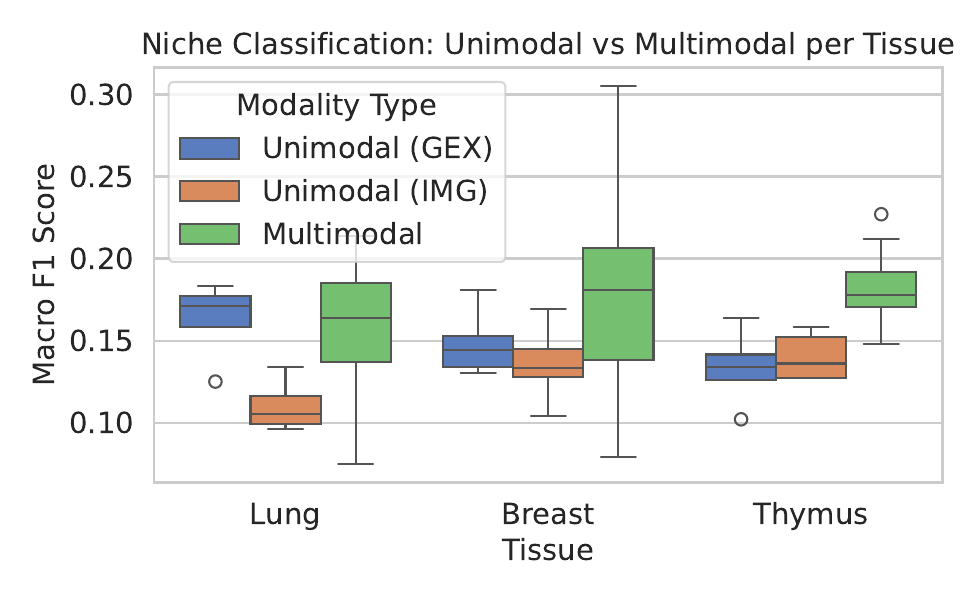} &
\includegraphics[width=0.3\textwidth]{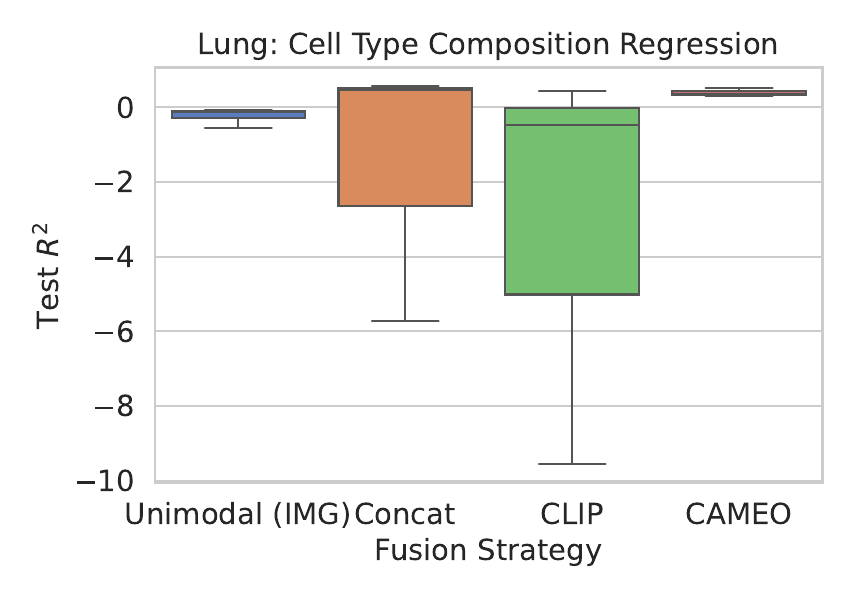} \\
\end{tabular}
\caption{Performance comparison by modality and fusion strategy. Left: Encoder combinations used in each fusion method across tissues. Center: Macro F1 scores for unimodal versus multimodal models using 100\% paired data. Right: Regression test $R^2$ for cell type composition prediction, showing multimodal models outperform unimodal approaches, with CAMEO achieving the highest performance.}
\label{fig:performance_comparison}
\end{center}
\vskip -0.2in
\end{figure}

% NEW RESULTS SECTION DRAFT FROM TILL:
\subsection{Results}
\label{exp_results}
% \subsubsection{Niche Classification}

All models were evaluated using the same train, validation, and test splits (stratified by patient). Due to significant class imbalances (shown in Appendix~\ref{app_datasets}), we use the macro F1 score as our primary metric for niche classification, and the $R^2$ score for cell type composition regression.

\begin{wrapfigure}{r}{0.45\textwidth}
\vskip -0.2in
\centering
\begin{minipage}{\linewidth}
    \centering
    % Figure
    \includegraphics[width=\linewidth]{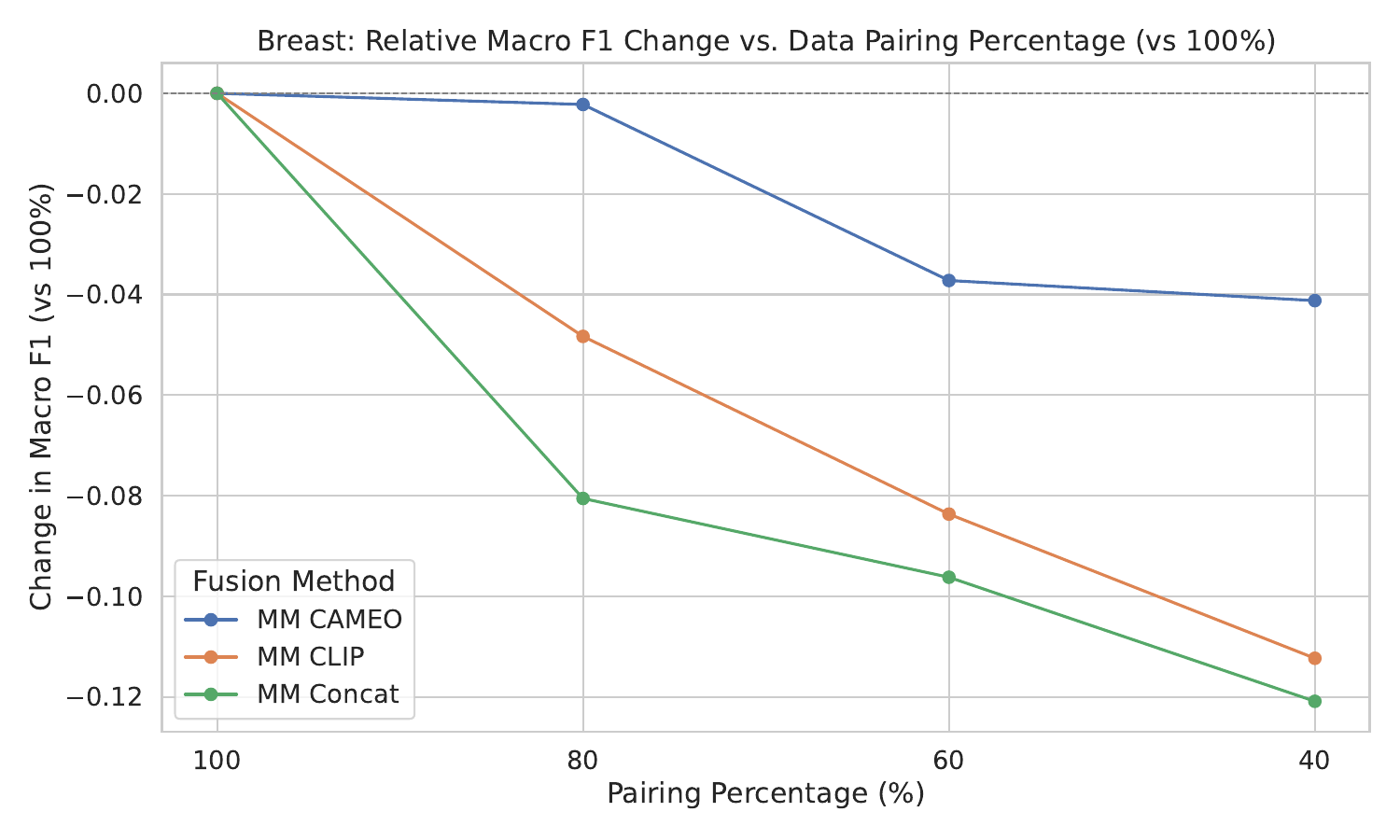}
    \captionof{figure}{Relative change in Breast macro F1 scores with decreasing pairing percentage. CAMEO shows greater robustness compared to CLIP and Concat. A similar patterns for Lung can be seen in App. Fig.~\ref{fig:app_pairing_other}.}
    \label{fig:pairing_robustness}
    \vspace{0.5em} % small spacing between figure and table

    %Table
    \captionof{table}{Niche classification performance under reduced data pairing scenarios using default UNI and scVI encoders.}
    \label{tab:reduced_pairing}
    \scriptsize
    \setlength{\tabcolsep}{3pt}
    \renewcommand{\arraystretch}{0.9}
    \begin{tabular}{lccccc}
    \toprule
    \textbf{Tissue} & \textbf{Method} & \textbf{100\%} & \textbf{80\%} & \textbf{60\%} & \textbf{40\% ($\Delta$)} \\
    \midrule
    \multirow{3}{*}{Lung}
    & CAMEO & 0.137 & \textbf{0.153} & \textbf{0.158} & \textbf{0.143} ($\Delta$ \textbf{4.4\%}) \\
    & CLIP & \textbf{0.194} & 0.146 & 0.128 & 0.133 ($\Delta$ -31.5\%) \\
    & Concat & 0.126 & 0.125 & 0.117 & 0.123  ($\Delta$ -2.2\%) \\
    \midrule
    \multirow{3}{*}{Breast}
    & CAMEO & \textbf{0.181} & \textbf{0.157} & 0.086 & \textbf{0.103}  ($\Delta$ -43.3\%) \\
    & CLIP & 0.170 & 0.103 & 0.108 & 0.078  ($\Delta$ -54.4\%) \\
    & Concat & 0.136 & 0.119 & \textbf{0.122} & 0.102 ($\Delta$ \textbf{-25.1\%}) \\
    \midrule
    % \multirow{3}{*}{Thymus}
    % & CAMEO & 0.185 & 0.187 & 0.101 & 0.115 ($\Delta$ -37.7\%) \\
    % & CLIP & \textbf{0.188} & \textbf{0.206} & \textbf{0.204} & \textbf{0.208} ($\Delta$ \textbf{10.6\%}) \\
    % & Concat & 0.152 & 0.166 & 0.151 & 0.143 ($\Delta$ -6.1\%) \\
    \bottomrule
    \end{tabular}
\end{minipage}
\vskip -0.2in
\end{wrapfigure}

Our experiments across three tissue types (Lung, Breast, and Thymus) yielded several consistent findings. First, multimodal approaches outperform their unimodal counterparts, demonstrating the value of integrating complementary information from images and gene expression data, as shown in Fig.~\ref{fig:performance_comparison} (center). 
This advantage is further validated in our cell type composition prediction task, where CAMEO achieves the highest $R^2$ scores (Fig.~\ref{fig:performance_comparison}, right). However, the performance gains vary across tissue types and encoder combinations, as illustrated in Fig.~\ref{fig:performance_comparison} (left). 
% \textcolor{red}{In the Lung dataset, Concat with CTransPath-GAT encoders achieved the highest macro F1 (0.214), while in Breast tissue, Concat with CONCH-GAT encoders performed best (0.305). For Thymus, CAMEO with CONCH-scVI yielded the top result (0.212).} 
In the Lung dataset, Concat–CTransPath–GAT achieved the highest macro F1 score (0.214). For Breast tissue, Concat–CONCH–GAT performed best (0.305), while in Thymus, Concat–CTransPath–scVI yielded the top result (0.227). Interestingly, CAMEO outperformed all Concat variants in terms of micro F1 score for both Breast and Thymus, achieving 0.445 with CAMEO–CONCH–scVI in Breast and 0.347 with CAMEO–CTransPath–Nicheformer in Thymus (Table~\ref{tab:model_comparison}). While Concat models are surprisingly strong when trained on 100\% paired data (in terms of macro F1), their performance declines as the proportion of paired data decreases, as we show next.

The relationship between pairing percentage and performance reveals that while performance typically decreases with less paired data, the rate of decline varies significantly between approaches. Notably, CAMEO demonstrates greater robustness to reduced pairing percentages than CLIP, particularly in the Breast dataset (Fig.~\ref{fig:pairing_robustness}). % \textcolor{red}{Similar trends are observed in the Lung dataset, while Thymus shows some unique patterns (see App Fig.~\ref{fig:app_pairing_other}). } 
Similar trends are observed in the Lung dataset (App. Fig.~\ref{fig:app_pairing_other}). This robustness pattern is consistent across both classification and regression tasks, with CAMEO maintaining stronger performance as paired data decreases (App. Fig.~\ref{fig:app_pairing_other}). These trends reflect mean performance across encoder combinations. Table~\ref{tab:reduced_pairing} provides detailed results using the UNI and scVI encoders: for the lung dataset, CAMEO shows a 4.4\% improvement at 40\% pairing compared to 31.5\% and 2.2\% drops for CLIP and Concat, respectively; for Breast tissue, CAMEO maintains the highest performance (0.103) despite reduced pairing.
% for Thymus tissue, CLIP shows better robustness (10.6\% improvement), suggesting tissue-specific strategies. These trends reflect mean performance across encoder combinations 
We attribute CAMEO's robustness to its adversarial alignment procedure, which effectively leverages unpaired data to preserve representation quality.

\subsection{Ablation Studies}
\subsubsection{Encoder Selection Impact Varies Across Frameworks}
% \begin{wraptable}{r}{0.45\textwidth}
% \begingroup
% \setlength{\tabcolsep}{3pt}
% \renewcommand{\arraystretch}{0.9}
% \scriptsize
% \centering
% \caption{Niche classification performance under reduced data pairing scenarios using default UNI and scVI encoders.}
% \label{tab:reduced_pairing}
% \begin{tabular}{lccccc}
% \toprule
% \textbf{Tissue} & \textbf{Method} & \textbf{100\%} & \textbf{80\%} & \textbf{60\%} & \textbf{40\% ($\Delta$)} \\
% \midrule
% \multirow{3}{*}{Lung}
% & CAMEO & 0.137 & \textbf{0.153} & \textbf{0.158} & \textbf{0.143} ($\Delta$ \textbf{4.4\%}) \\
% & CLIP & \textbf{0.194} & 0.146 & 0.128 & 0.133 ($\Delta$ -31.5\%) \\
% & Concat & 0.126 & 0.125 & 0.117 & 0.123  ($\Delta$ -2.2\%) \\
% \midrule
% \multirow{3}{*}{Breast}
% & CAMEO & \textbf{0.181} & \textbf{0.157} & 0.086 & \textbf{0.103}  ($\Delta$ -43.3\%) \\
% & CLIP & 0.170 & 0.103 & 0.108 & 0.078  ($\Delta$ -54.4\%) \\
% & Concat & 0.136 & 0.119 & \textbf{0.122} & 0.102 ($\Delta$ \textbf{-25.1\%}) \\
% \midrule
% % \multirow{3}{*}{Thymus}
% % & CAMEO & 0.185 & 0.187 & 0.101 & 0.115 ($\Delta$ -37.7\%) \\
% % & CLIP & \textbf{0.188} & \textbf{0.206} & \textbf{0.204} & \textbf{0.208} ($\Delta$ \textbf{10.6\%}) \\
% % & Concat & 0.152 & 0.166 & 0.151 & 0.143 ($\Delta$ -6.1\%) \\
% \bottomrule
% \end{tabular}
% \endgroup
% \end{wraptable}

Our results reveal that encoder selection critically impacts performance, but optimal combinations differ across multimodal frameworks. As an example, in the Lung dataset, the best-performing CAMEO variant is CAMEO–UNI–Nicheformer (0.193), whereas Concat achieves its highest score with Concat–CTransPath–GAT (0.214). This inconsistency extends to other tissues as well, where the best-performing encoder pair for one framework rarely aligns with the optimal choice for another (Table~\ref{tab:model_comparison}). This finding underscores the importance of comprehensive encoder evaluations when implementing any multimodal approach.

\subsubsection{Unimodal Performance Does Not Predict Multimodal Success}
\begin{wrapfigure}{r}{0.45\textwidth}
% \vspace{-15pt}
\vskip -0.55in
\includegraphics[width=\linewidth]{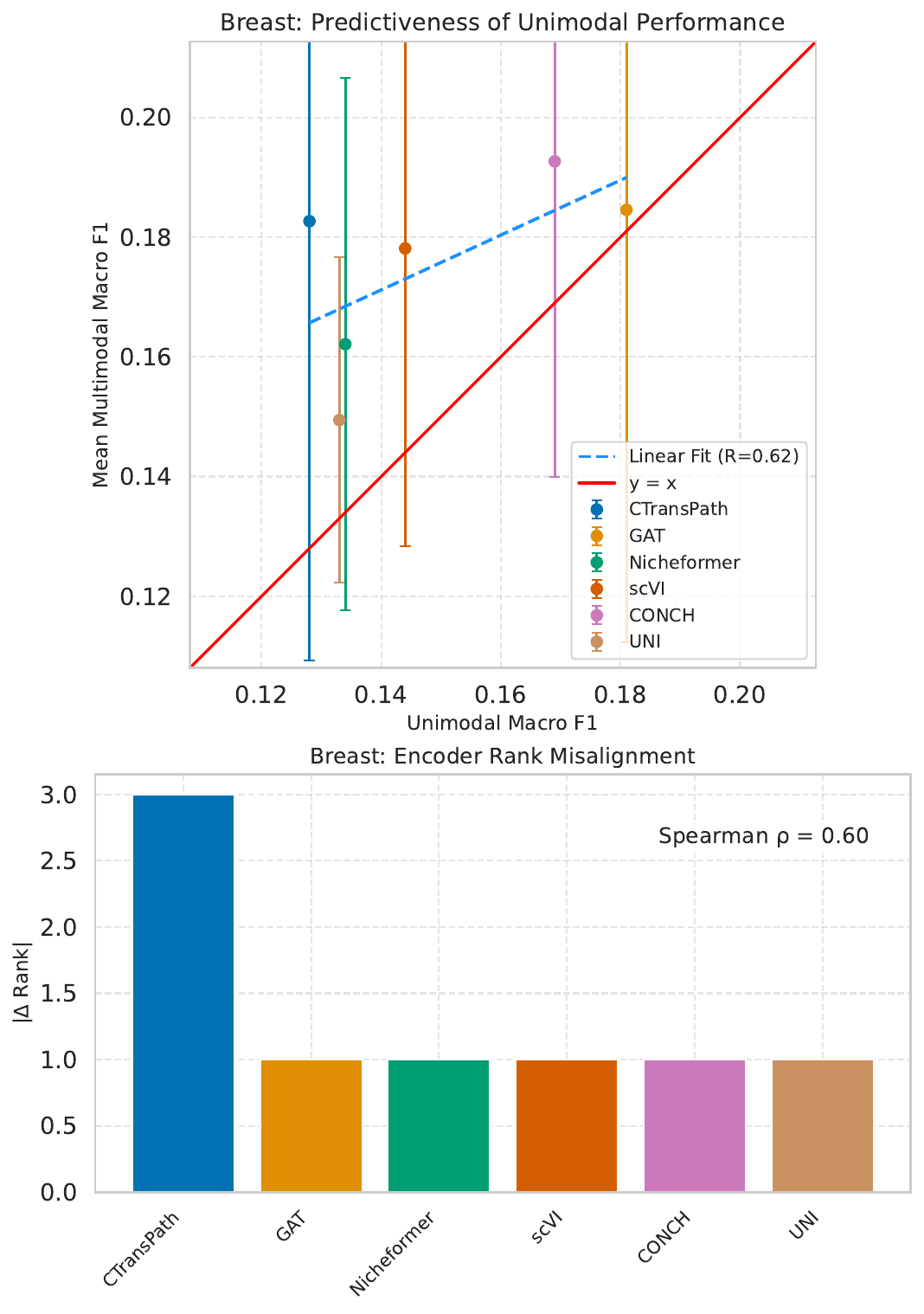}
% \vspace{-20pt}
{\small
\caption{Predictiveness of unimodal performance for multimodal success in the Breast dataset. Moderate correlation (Spearman $\rho = 0.60$) and high variance across multimodal models, and substantial rank misalignments (bottom panel). Results for Lung and Thymus are shown in App. Fig.~\ref{fig:app_predictiveness}.}
\label{fig:predictiveness}
}
\vskip -0.55in
\end{wrapfigure}

A surprising finding is that the performance of a unimodal encoder poorly predicts its effectiveness in multimodal settings, as illustrated in Fig.~\ref{fig:predictiveness} for the Breast dataset. For example, while GAT achieves the highest macro F1 score (0.181) among gene expression encoders for this dataset, the best-performing multimodal combination in CLIP and CAMEO uses Nicheformer instead. This pattern persists across tissues (App. Fig.~\ref{fig:app_predictiveness}), suggesting that complementarity between encoders may be more important than individual encoder strength. This non-predictive relationship challenges the common practice of selecting encoders based solely on their standalone performance and highlights the need for systematic evaluation of encoder combinations in multimodal frameworks.

\subsection{Recommendations}
 Based on our findings, we recommend the following best practices for developing multimodal frameworks: (i) systematically evaluate multiple encoder combinations, as the optimal pairing of gene expression and image encoders can vary across both tissue types and multimodal integration methods; (ii) include concatenation as a competitive baseline, since it often performs well; and (iii) consider adversarial methods like CAMEO when working with limited paired data or in settings with label imbalance, as they can effectively exploit unpaired examples.

\section{Discussion}
\label{discussion}

Our study introduces adversarial alignment as an effective alternative to existing multimodal fusion approaches in biomedical settings where paired samples are scarce. CAMEO's superior robustness to reduced paired data demonstrates the value of adversarial mechanisms for leveraging unpaired examples, which is particularly valuable in computational pathology, where unimodal data is abundant but generating paired datasets remains challenging. Unexpectedly, we found that the performance of unimodal encoders is a poor predictor of multimodal success, challenging conventional wisdom in model selection and highlighting the importance of encoder complementarity. This suggests that the way encoders interact within a fusion framework may be more important than their individual strengths, requiring evaluation of multiple combinations when implementing multimodal approaches. While our work focused on spatial omics and histopathology, the adversarial alignment principles explored here could extend to other domains with limited paired data. By releasing expert-annotated multimodal datasets and providing practical insights for modality fusion strategies, we contribute both methodological advancement and resources to facilitate further exploration of multimodal learning in data-constrained environments.

\ificlrfinal
\subsubsection*{Acknowledgments}
K.T., T.R. and R.H.G. are supported by the Helmholtz Association under the joint
research school ``Munich School for Data Science -- MUDS''. T.R. is additionally
supported by the Munich Center for Machine Learning (MCML), funded by the Federal
Ministry of Research, Technology and Space (BMFTR) and the Free State of Bavaria.
\fi

\newpage

\bibliography{iclr2026_conference}
\bibliographystyle{iclr2026_conference}

\newpage

\appendix

\section{Data and Code Availability}
All data and code necessary to reproduce the reported results will be released upon publication.

\section{LLM Usage}
We used a large language model (LLM) to assist with text editing and figure formatting. The LLM was used only for improving clarity and presentation; all scientific content, analyses, and interpretations were generated by the authors.

\section{Limitations}
\label{app_limitations}

CAMEO builds on the hypothesis that two modalities can provide distinct but complementary information. This assumption introduces certain limitations:
\begin{itemize}
    \setlength{\itemsep}{2pt}  
    \setlength{\parskip}{1pt}
    \setlength{\parsep}{1pt}
    
    \item \textbf{Representation Quality. } CAMEO aligns representations learned by self-supervised unimodal encoders. Its ability to meaningfully align these representations for downstream tasks is limited by the quality of the unimodal representations. This limitation is an opportunity at the same time, as the modular design allows the user to replace the encoders with improved models.
    \item \textbf{Modality Translation. } The assumption of complementarity limits the alignment of modalities that do not share common concepts. If one modality, for example, is independent of the other, the translation becomes random. 
    \item \textbf{Downstream Task Suitability. } Our approach was tested on computational pathology tasks chosen for their inherently multimodal nature, where both histology and gene expression data are informative. This approach might not make sense for every task. Specifically, it may not generalize to tasks for which one modality yields little or no added benefit.
    \item \textbf{Loss of Modality-Specific Information. } The alignment process may reduce modality-specific information, as it prioritizes shared representations. A potential future direction is to learn both shared and modality-specific latent factors, preserving information unique to each modality while still benefiting from their alignment.
\end{itemize}

\section{Additional Results}
\label{app_results}

This section provides additional results and visualizations that complement the main findings presented in Section~\ref{exp_results}.

\subsection{Impact of Reduced Pairing Percentages}
\label{app_pairing}

\begin{figure*}[t]
\begin{center}
\begin{tabular}{cc}
\includegraphics[width=0.48\textwidth]{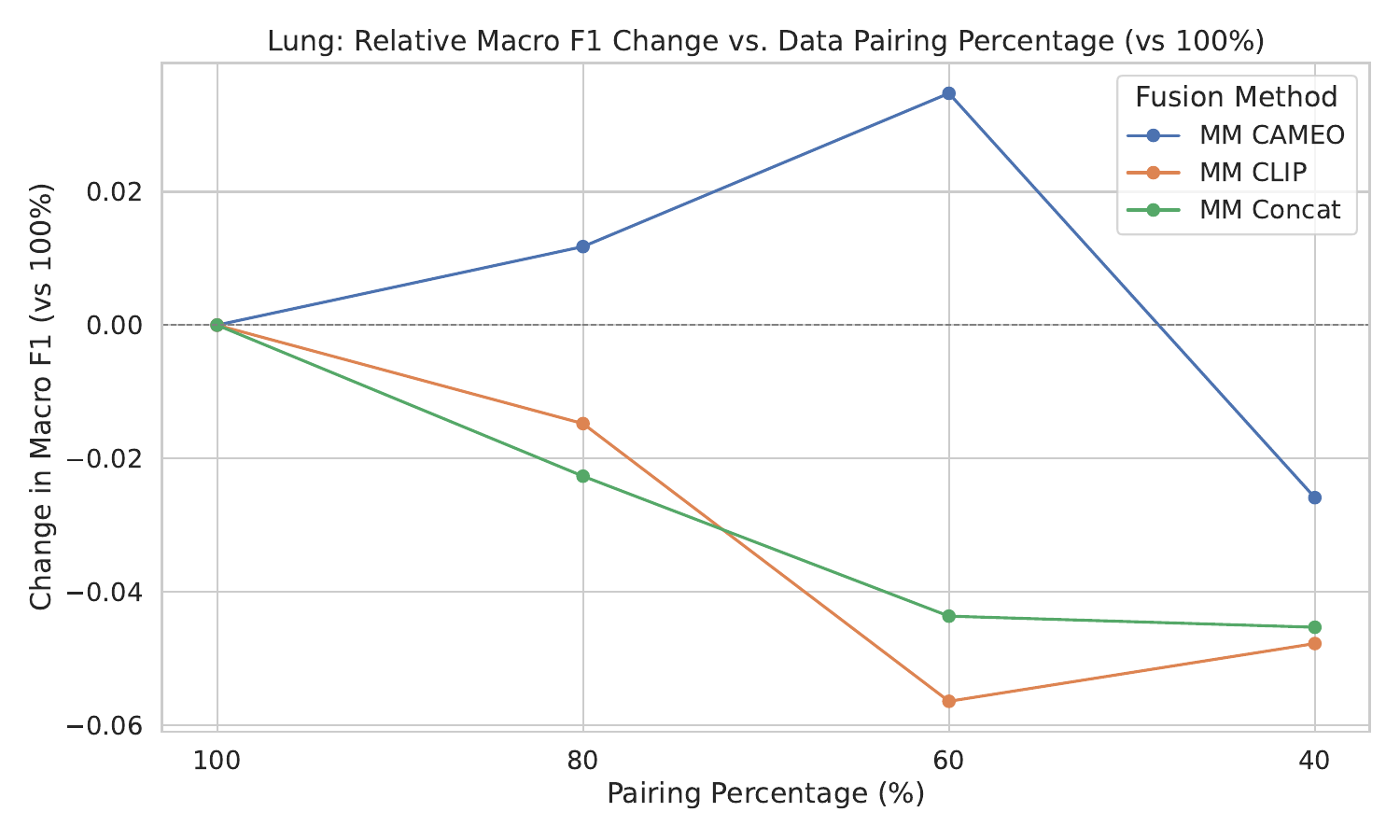} &
\includegraphics[width=0.48\textwidth]
{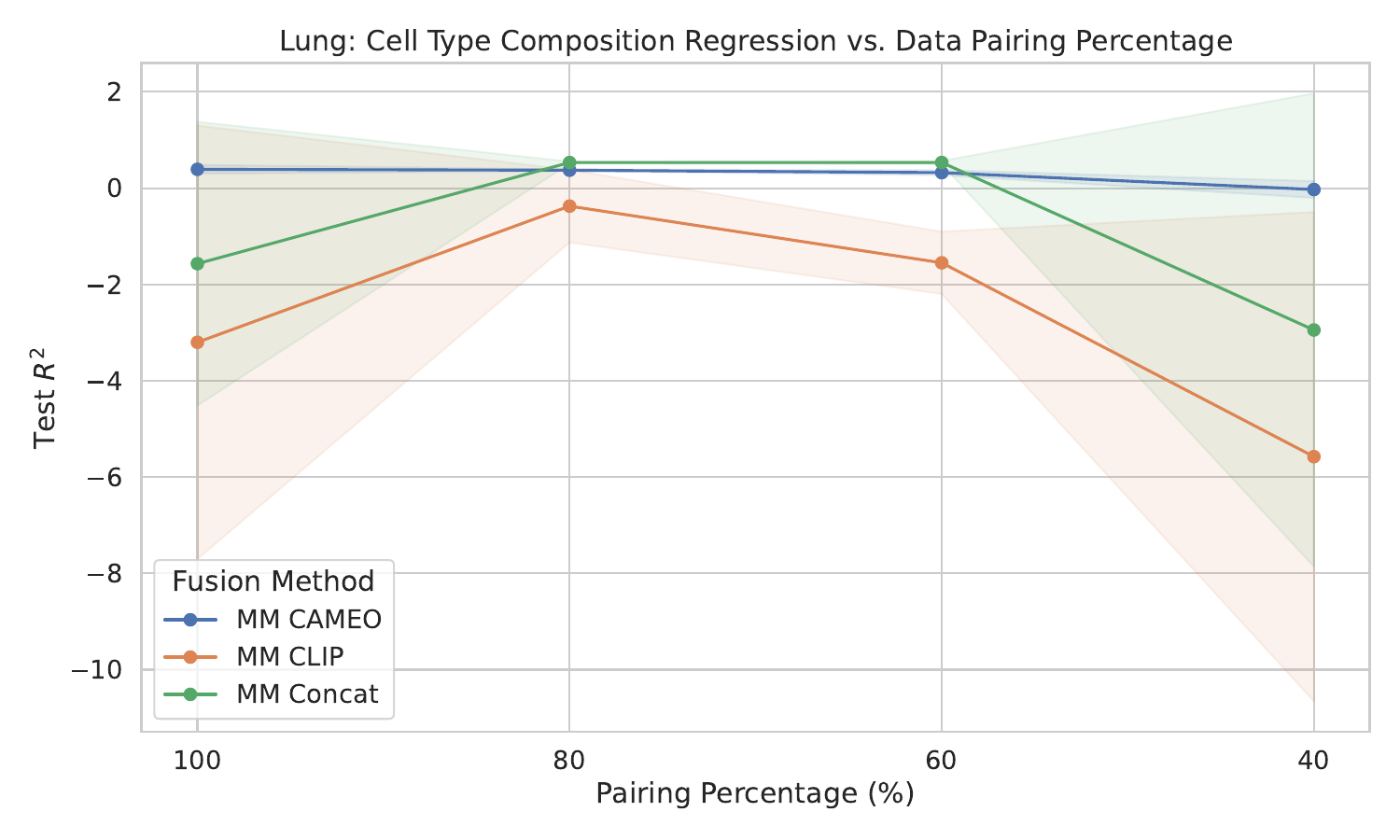} \\
% {chapters/figures/relative_change_macro_f1_pairing_percentage_thymus.pdf} \\
\end{tabular}
\caption{Lung dataset: Relative change in macro F1 scores compared to 100\% pairing (left). Lung shows similar patterns to Breast (Fig.~\ref{fig:pairing_robustness}). Mean and standard deviation of absolute $R^2$ scores for cell type composition prediction (right).}
\label{fig:app_pairing_other}
\end{center}
\end{figure*}

Fig.~\ref{fig:app_pairing_other} shows the relative performance changes for the Lung dataset as the percentage of paired data decreases, complementing the Breast dataset results presented in Fig.~\ref{fig:pairing_robustness}.
% Fig.~\ref{fig:app_mean_std} provides a complementary view by showing the absolute macro F1 scores for all three fusion approaches across all tissues.

% \begin{figure*}[t]
% \begin{center}
% \begin{tabular}{cc}
% \includegraphics[width=0.5\textwidth]{chapters/figures/mean_std_macro_f1_pairing_percentage_lung.pdf} &
% \includegraphics[width=0.5\textwidth]{chapters/figures/mean_std_macro_f1_pairing_percentage_breast.pdf} \\
% \includegraphics[width=0.5\textwidth]{chapters/figures/mean_std_macro_f1_pairing_percentage_thymus.pdf} &
% \includegraphics[width=0.5\textwidth]{chapters/figures/reg_test_r2_pairing_percentage_lung.pdf} \\
% \end{tabular}
% \caption{Mean and standard deviation of macro F1 scores across different pairing percentages for Lung (left), Breast (center), and Thymus (right) datasets. Shaded regions represent the standard deviation across different encoder combinations.}
% \label{fig:app_mean_std}
% \end{center}
% \end{figure*}

\subsection{Predictiveness of Unimodal Performance}
\label{app_predictiveness_section}

\begin{figure*}[t]
\begin{center}
\begin{tabular}{cc}
\includegraphics[width=0.48\textwidth]{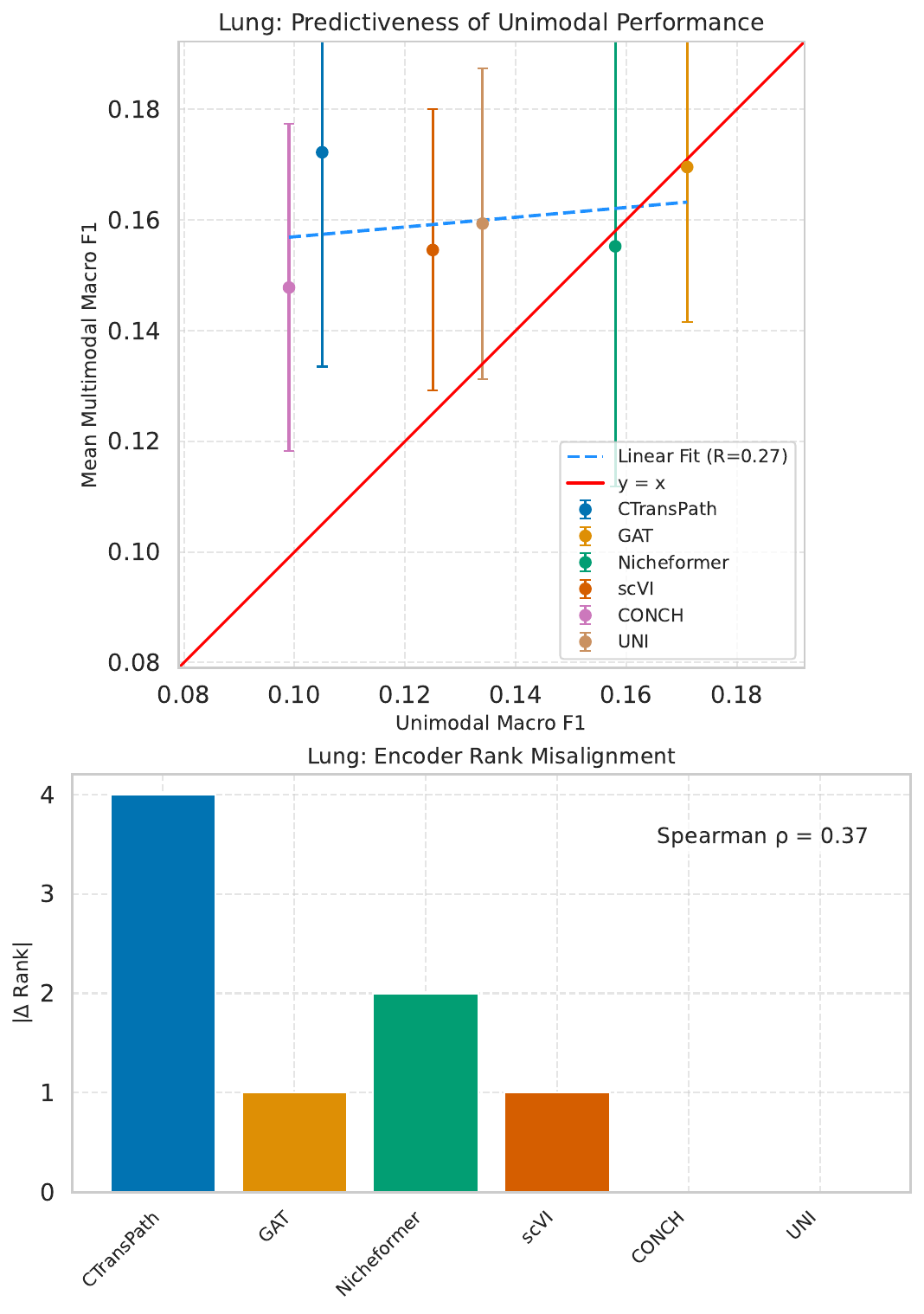} &
\includegraphics[width=0.48\textwidth]{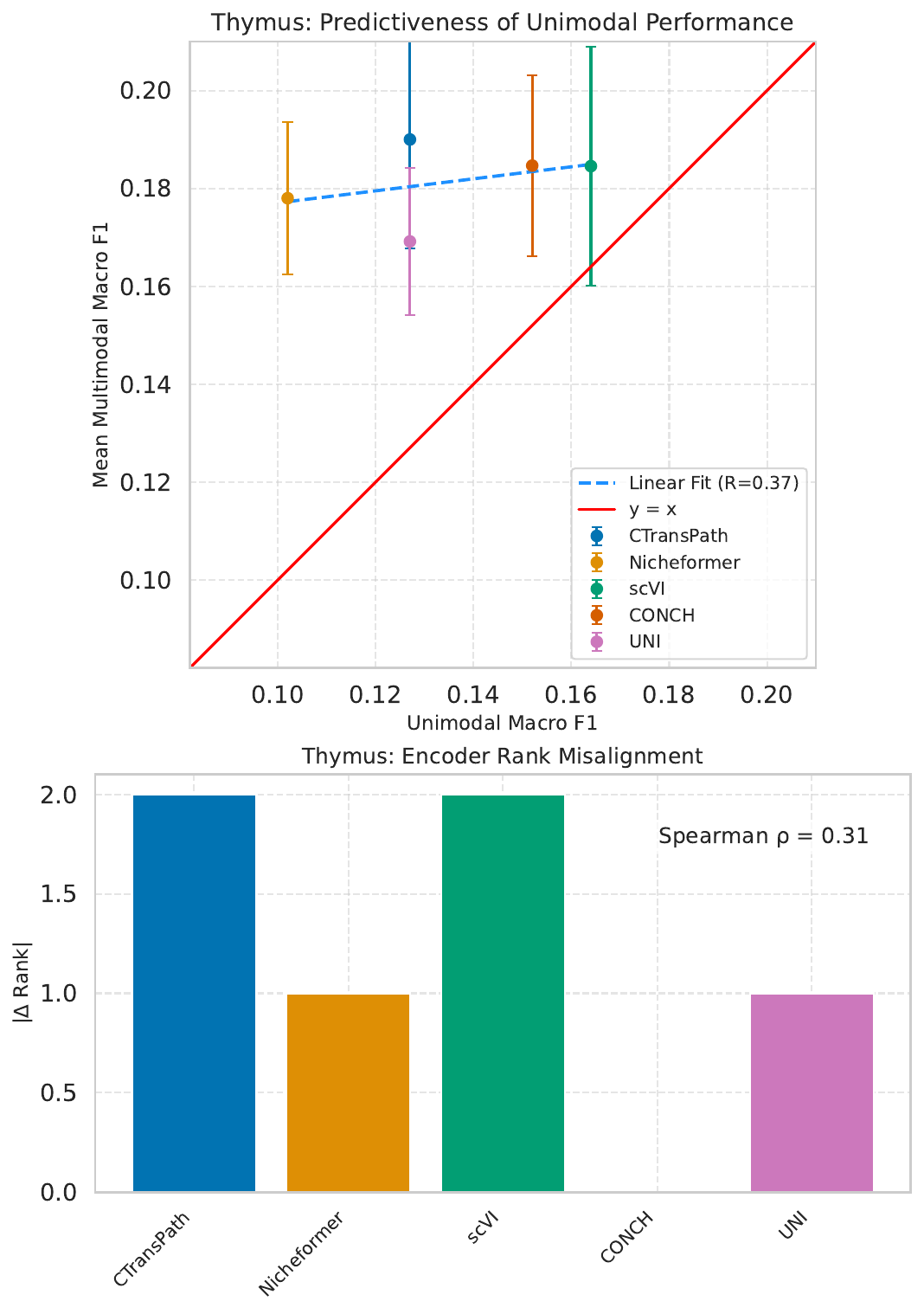} \\
\end{tabular}
\caption{Predictiveness of unimodal performance for multimodal success in Lung (left) and Thymus (right) datasets. Both tissues show weak correlations ($\rho = 0.37$ and $\rho = 0.31$, respectively) and significant rank misalignments, further supporting the finding that unimodal performance poorly predicts multimodal success.}
\label{fig:app_predictiveness}
\end{center}
\end{figure*}

Fig.~\ref{fig:app_predictiveness} extends the analysis presented in Fig.~\ref{fig:predictiveness} to the Lung and Thymus datasets. The weak correlations (Spearman $\rho$ values of 0.37 and 0.31, respectively) confirm that the pattern observed in the Breast dataset ($\rho = 0.60$) is consistent across tissues, though with varying strengths.

\section{Model Details}
\label{app_model_details}

\subsection{Image Baselines}
The image baselines leverage pretrained encoders to extract embeddings from histopathological images. These encoders are used in a frozen state, with a task-specific head trained on top.
\begin{itemize}
    \setlength{\itemsep}{2pt}  
    \setlength{\parskip}{1pt}
    \setlength{\parsep}{1pt}
    \item \textbf{UNI}~\citep{uni}: UNI is built on a ViT-Large (ViT-L) architecture and pretrained on over 100 million image patches derived from 100,000 H\&E-stained whole-slide images (WSIs) spanning 20 major tissue types.
    \item \textbf{CTransPath}~\citep{ctrans}: Integrates a convolutional neural network with a multi-scale Swin Transformer architecture. It is pretrained on 15 million unlabeled image patches from The Cancer Genome Atlas (TCGA) and PAIP datasets.
    \item \textbf{ResNet-50}~\citep{resnet}: A convolutional neural network pre-trained on ImageNet-1k.
    \item \textbf{Virchow}~\citep{virchow}: A self-supervised vision transformer pretrained on 3.1 million whole-slide histopathology images.
    \item \textbf{CONCH}~\citep{conch}: A vision-language model trained via contrastive learning on 1.17 million histopathology-specific image–caption pairs.
\end{itemize}

% \paragraph{Vision Encoder}
% We use a pre-trained ViT-L/16 variant of the Vision Transformer~\citep{vit16x16} as our vision encoder. The model comprises $24$ Transformer layers, each with 16 attention heads and a hidden size of $1024$, resulting in a total of $\sim307M$ parameters. Multi-head self-attention and feed-forward blocks operate on the patch embeddings to capture global contextual information. This ViT, called UNI, is trained using the DINOv2~\citep{dinov2} framework. The model processes each niche patch as input and uses the final $CLS$ token representation from this encoder for downstream tasks.

\subsection{Gene Expression Baselines}
Gene expression baselines process high-dimensional omics data, where each sample is represented as a vector of gene expression values.
\begin{itemize}

    \item \textbf{GAT}: Graph Attention Networks (GATs)~\citep{velickovic2017graph} learn node representations by dynamically adjusting the influence of neighboring nodes via attention weights. The attention mechanism computes a weighted aggregation of neighbors' features, where the attention coefficients are learned based on the similarity of node features. This allows the model to focus on more relevant neighbors, making GATs particularly suitable for graph-structured data such as spatial transcriptomics measurements. In this context, $\alpha_{uv}$ represents the attention weight between nodes $u$ and $v$, which is learned as a function of the features of both nodes, modulated by the graph structure $G$.
    While GATs are typically trained in an unsupervised manner, self-supervised learning (SSL) has been shown to enhance their performance, particularly when applied to large datasets~\citep{richter2023spatialssl}. The model is pre-trained using masked modeling, where parts of the gene expression profile $x_v$ are masked, and the objective is to predict the masked entries from the unmasked ones. Let $f_\theta(x_v)$ denote the unimodal encoder, which maps the gene expression vector $x_v$ to an embedding $z_v$ used later in the fusion module. The training objective is to minimize the masked modeling loss, which for a set of masked nodes $\mathcal{M}$ is:
    \begin{equation}
    \mathcal{L}_{\text{SSL}} = \sum_{v \in \mathcal{M}} || (1_{p} - M_v) \odot f_\theta(M_v \odot x_v) - \hat{x}_v ||_2^2,
    \end{equation}
    where $M_v \in \{0, 1\}^p$ is a mask for the genes of cell $v$ and $(1_{p} - M_v)$ ensures that only the masked entries contribute to the loss. This SSL pretraining enables the model to learn meaningful representations from large unimodal datasets without requiring labeled data. By capturing complex spatial dependencies in gene expression patterns, the learned embeddings provide a strong foundation for downstream tasks.
    
    \item \textbf{scVI}~\citep{scvi}: A probabilistic framework for analyzing gene expression data. It employs a variational autoencoder (VAE) to learn a low-dimensional latent representation of gene expression profiles.
    \item \textbf{scGPT}~\citep{cui2024scgpt}: Adapts the GPT architecture for gene expression data by treating genes as words (tokens) and cells as sentences. It is pretrained on 33 million cross-tissue human cells.
    \item \textbf{CellPLM}~\citep{cellplm}: A spatial transcriptomics foundation model that explicitly encodes cell spatial relationships. CellPLM is trained on 9 million scRNA-seq cells and 2 million spatial transcriptomics cells. % A foundation model designed for single-cell gene expression data. It treats cells as tokens, leveraging 9 million scRNA-seq cells and 2 million spatial transcriptomics cells for pretraining.
    \item \textbf{Nicheformer}~\citep{nicheformer}: A transformer-based foundation model tailored for spatial single-cell omics. It is pretrained on 57 million dissociated and 53 million spatially resolved cells from 73 human and mouse tissues.
\end{itemize}

% \paragraph{GAT}
% We implement a Graph Attention Network (GAT) using PyTorch Geometric's GATConv layers. The architecture consists of:
% \begin{itemize}
%     \setlength{\itemsep}{2pt}  
%     \setlength{\parskip}{1pt}
%     \setlength{\parsep}{1pt}
%     \item Input dimension: 343 genes (lung) or 313 genes (breast)
%     \item Hidden dimension: 128
%     \item 3 GAT layers with ReLU activation
%     \item Optional negative binomial prediction head (default: True)
%     \item Patch-level attention pooling layer
% \end{itemize}

% The model processes spatial transcriptomics data as graphs, where each cell represents a node with gene expression values as features. We construct patches from the spatial data using the following protocol:
% \begin{itemize}
%     \setlength{\itemsep}{2pt}  
%     \setlength{\parskip}{1pt}
%     \setlength{\parsep}{1pt}
%     \item Cells are grouped into patches based on spatial coordinates. These patches are equivalent to the patches used by the vision encoder
%     \item Edge connections are established between cells within a defined radius of $20\mu m$
%     \item Each patch results in one full graph
%     \item Patches with too few cells ($<20$) are removed
% \end{itemize}

\subsubsection{Multimodal Baselines}
\begin{itemize}
    \setlength{\itemsep}{2pt}  
    \setlength{\parskip}{1pt}
    \setlength{\parsep}{1pt}
    \item \textbf{CLIP} \citep{clip}: A multimodal model trained via contrastive learning. Each modality is independently encoded and projected into modality-specific latent spaces. A contrastive loss aligns paired samples, enabling joint representation learning.
    \item \textbf{Concatenated}: Concatenates the embeddings produced by the gene expression and image encoders, without any explicit alignment or fusion mechanism.
    % \textcolor{red}{\item \textbf{UIMC}~\citep{uimc}:}
    % \textcolor{red}{\item \textbf{MM-LEGO}~\citep{mm-lego}:}
\end{itemize}

\subsection{CAMEO Architecture}
\begin{itemize}
    \item \textbf{Input Dimensions:}
    \begin{itemize}
        \item Image embeddings: $d_{\text{img}} = 1024 \ \text{(UNI)} \ \text{or} \ 768 \ \text{(CTransPath)} \ \text{or} \ 512 \ \text{(CONCH)}$
        \item Gene expression embeddings: $d_{\text{gex}} = 128 \ \text{(GAT)} \ \text{or} \ 128 \ \text{(scVI)} \ \text{or} \ 512 \ \text{(Nicheformer)}$
        \item Common latent space: $d=\min(d_{img}, d_{gex})$
    \end{itemize}
    
    \item \textbf{Generator Architecture:}
    \begin{itemize}
        \item Encoder layers: $d \rightarrow d \rightarrow d/2$
        \item Decoder layers: $d/2 \rightarrow d \rightarrow d$
        \item Shared encoder layer: $d/2 \rightarrow d/4$
        \item Shared decoder layer: $d/4 \rightarrow d/2$
        \item All layers include BatchNorm1d and LeakyReLU ($\alpha=0.2$)
    \end{itemize}
    
    \item \textbf{Discriminator Architecture:}
    \begin{itemize}
        \item Input layer: $d \rightarrow d/4$
        \item Hidden layer: BatchNorm1d + LeakyReLU
        \item Output layer: $d/4 \rightarrow 1$
    \end{itemize}
\end{itemize}

\subsection{CAMEO Training Configuration}
\begin{itemize}
    \item Shared layers use LeakyReLU activation ($\alpha=0.2$)
    \item Batch normalization in discriminators
    \item Weight clipping range: $[-0.01, 0.01]$ for Wasserstein GAN
    \item Optimizers: RMSprop (Wasserstein) or AdamW (standard GAN)
\end{itemize}

\subsection{Training Protocols}
\label{app_training_protocols}

\paragraph{Gene Expression Encoder Pre-training}
The GAT model is pre-trained using a masked gene expression prediction task:
\begin{itemize}
    \setlength{\itemsep}{2pt}  
    \setlength{\parskip}{1pt}
    \setlength{\parsep}{1pt}
    \item Masking ratio: 0.4 (40\% of expression values masked)
    \item Learning rate: 0.001
    \item Batch size: 1 (full patch per batch)
    \item Early stopping patience: 200 epochs
    \item Maximum epochs: 5000
    \item Gradient clipping: 1.0
    \item Loss: MSE or Negative Binomial (when predict\_nb=True, Negative Binomial performs better)
\end{itemize}

\paragraph{scVI Training} The scVI models for the Lung and Breast datasets were pre-trained using default parameters from the scVI-tools pipeline. For Thymus, the model was pre-trained following the authors’ recommendations \citep{thymus}, using sample ID as a batch covariate.

\paragraph{CLIP and CAMEO Training}
For CLIP and CAMEO models, we evaluated a pre-training duration of 10 epochs. For CAMEO specifically, we optimized $\lambda_{align}$ to balance the contribution of the alignment loss term.

\paragraph{Linear Probing Stage}
We adopted a linear probing setup to isolate the quality of the pretrained representations without updating encoder weights. Linear probes were trained using the AdamW optimizer with a learning rate of 1e-3 for 10 epochs. This learning rate was found to perform robustly across different model initializations and tasks.

\paragraph{CAMEO-specific Parameters}
\begin{itemize}
    \setlength{\itemsep}{2pt}  
    \setlength{\parskip}{1pt}
    \setlength{\parsep}{1pt}
    \item $\lambda_{align}$: Controls the weight of alignment loss
    \item Generator Learning Rate: 1e-4
    \item Discriminator Learning Rate: 5e-5
    \item Weight Clipping Value: 0.01
\end{itemize}

\section{Evaluation Metrics}
\label{app_metrics}

Let $y \in \{1,\dots,C\}$ be the true class label and $\hat{y}$ the predicted class label for $N$ samples.

\paragraph{Macro F1 score.}  
The F1 score for class $c$ is defined as
\[
\mathrm{F1}_c = \frac{2\, \mathrm{Precision}_c \cdot \mathrm{Recall}_c}{\mathrm{Precision}_c + \mathrm{Recall}_c},
\]
where
\[
\mathrm{Precision}_c = \frac{\mathrm{TP}_c}{\mathrm{TP}_c + \mathrm{FP}_c}, \quad
\mathrm{Recall}_c = \frac{\mathrm{TP}_c}{\mathrm{TP}_c + \mathrm{FN}_c}.
\]
The macro F1 score averages $\mathrm{F1}_c$ over all $C$ classes with equal weight:
\[
\mathrm{MacroF1} = \frac{1}{C} \sum_{c=1}^C \mathrm{F1}_c.
\]
Macro F1 treats all classes equally and is sensitive to performance on underrepresented classes.

\paragraph{Micro F1 score.}  
The micro F1 score computes precision and recall by first aggregating $\mathrm{TP}$, $\mathrm{FP}$, and $\mathrm{FN}$ across all classes, and then applying the F1 formula to these aggregated counts. Micro F1 reflects overall model performance weighted by class frequency.

\paragraph{$R^2$ score.}  
For regression tasks, the coefficient of determination is defined as
\[
R^2 = 1 - \frac{\sum_{i=1}^N (y_i - \hat{y}_i)^2}{\sum_{i=1}^N (y_i - \bar{y})^2},
\]
where $\bar{y}$ is the sample mean of $\{y_i\}_{i=1}^N$. $R^2$ measures the proportion of variance explained by the model.

\section{Comparison to Probabilistic Alignment Methods
\label{app_prob_align}}

We compare two prominent probabilistic alignment methods in the context of histopathology-gene expression alignment:

% Discussed in the Related Works section
% \textbf{SCOT}~\citep{demetci2022scot} uses Gromov-Wasserstein optimal transport to align multi-omics data by minimizing the distortion between intra-modality distances. The method excels at fully unpaired alignment but requires $O(n^2)$ memory for the all-pairs shortest path computation, limiting its applicability to large-scale datasets.

\textbf{DDMEC}~\citep{bounoua2025learning} learns stochastic conditionals between modalities using diffusion models with KL-regularized ``soft marginals''. The approach offers principled uncertainty modeling but requires significant computational resources for training and sampling from the diffusion models.

\textbf{scTopoGAN}~\citep{singh2023sctopogan} combines topology-preserving autoencoders with GAN-based latent alignment. The method assumes shared topological structure between modalities - an assumption well-suited for molecular measurements (e.g., ATAC-seq, CITE-seq) but less applicable when modalities capture fundamentally different biological features at distinct scales.

For histopathology-gene expression alignment, key practical considerations include:
\begin{enumerate}
    \item Computational efficiency for large spatial patches
    \item Non-coinciding manifold structure between modalities
    \item Availability of paired anchors, even if limited
    \item Requirements for stable embeddings in downstream tasks
\end{enumerate}

These constraints favor deterministic approaches like CAMEO for this specific use case, though probabilistic methods remain valuable in settings where their assumptions hold and computational constraints permit.

\section{Computational Resources}
\label{app_comp_res}
% \textcolor{red}{TODO: Check the latest specs.} The GAT was trained using Python 3.10 on a GPU server with these specifications:

% \begin{itemize}
%     \item GPU: 16 Tesla V100 GPUs, each with 32GB of RAM
%     \item GPU: 2 Tesla V100 GPUs, each with 16GB of RAM
%     \item GPU: 8 A100-SXM4 GPUs, each with 40GB of RAM
% \end{itemize}

% CAMEO was trained using Python 3.10 on a GPU server with a single A100-SXM4 GPU (40GB RAM).

Models were trained using Python 3.10 on a GPU server with these specifications:

\begin{itemize}
    \setlength{\itemsep}{2pt}  
    \setlength{\parskip}{1pt}
    \setlength{\parsep}{1pt}
    \item GPU: 30 nodes, each with 96 CPUs, 768GB memory, 4 NVIDIA H100 GPUs each with 94GB HBM2 memory.
    \item GPU: 5 nodes, each with 96 CPUs, 1TB memory, 4 NVIDIA A100 GPUs each with 80GB HBM2 memory.
    \item GPU: 4 nodes, each with 252 CPUs, 2TB memory, 8 NVIDIA A100 GPUs each with 80GB HBM2 memory.
\end{itemize}

Training each model typically took less than 6 hours, and in most cases, less than 2 hours.

\section{Datasets}
\label{app_datasets}
% \subsection{Dataset generation}
\begin{figure*}[h]
\vskip 0.2in
\begin{center}
\includegraphics[width=\textwidth]{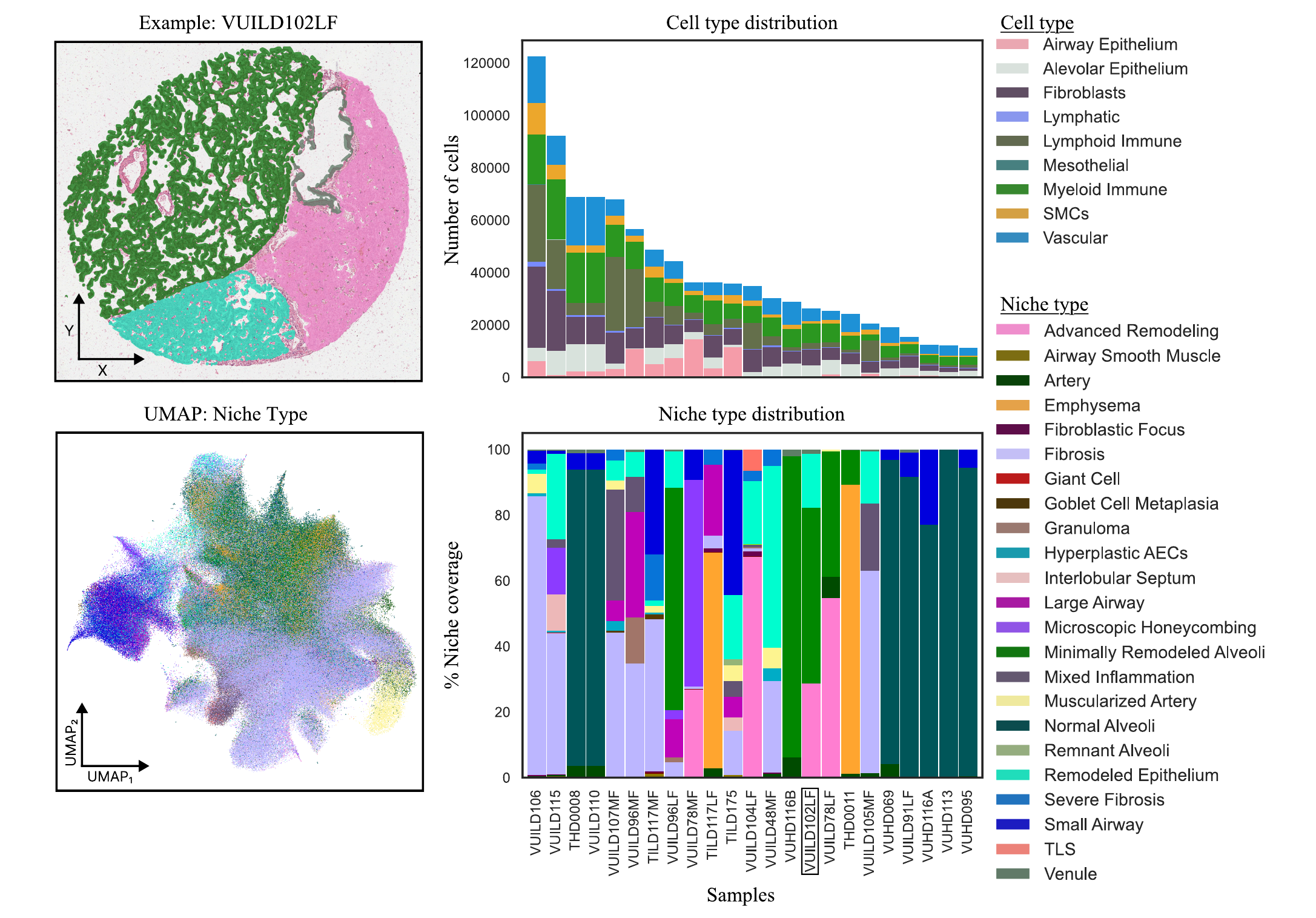}
\caption{Metadata overview for the LungPF Xenium cohort~\citep{lpf}, highlighting cell-type distributions, niche-label distributions, and the imbalance of labels across samples. Top left: example histology image with expert-provided niche annotations overlaid; the corresponding niche-label distribution for this sample is shown in the bar chart (sample name boxed for reference). Top right: total cell-type distribution across all 23 samples, showing slight variability in cell-type composition. Bottom: UMAP of patch-level gene expression profiles and overall niche-label distribution, revealing substantial imbalance across the 23 unique annotated niche types. Niches labeled `NOANNOT' are excluded from all visualizations.}
\label{fig:data_dist}
\end{center}
\vskip -0.2in
\end{figure*}

Paired spatial transcriptomics datasets remain scarce. Often, spatial datasets are released with only the genomic component and its corresponding spatial context, while the paired histology is omitted due to the prohibitive costs of publishing large images. While this is gradually improving, collecting large-scale paired datasets remains challenging. To support research on partially paired spatial data, we created a HuggingFace dataset that combines histology images and gene expression data at the niche level. A niche is defined as a $224\times224$ pixel crop of the histology slide, paired with the corresponding gene expression values from cells located within the crop. Additionally, expert-provided niche annotations are included for each pair, along with the total number of cells per spot, as estimated by Xenium’s default segmentation.

To ensure proper spatial alignment between the histological and transcriptomic data, we utilized SpatialData~\citep{spatialdata}. The niches are tessellated across each slide without overlap, and we apply a quality control filter to exclude niches containing less than 50\% tissue (i.e., those that predominantly contain background). For simplicity and interpretability, transcripts from partially included cells are treated as whole-cell data within the niche.

\begin{figure*}[h]
\vskip 0.2in
\centering

\begin{subfigure}{\textwidth}
    \includegraphics[width=\textwidth]{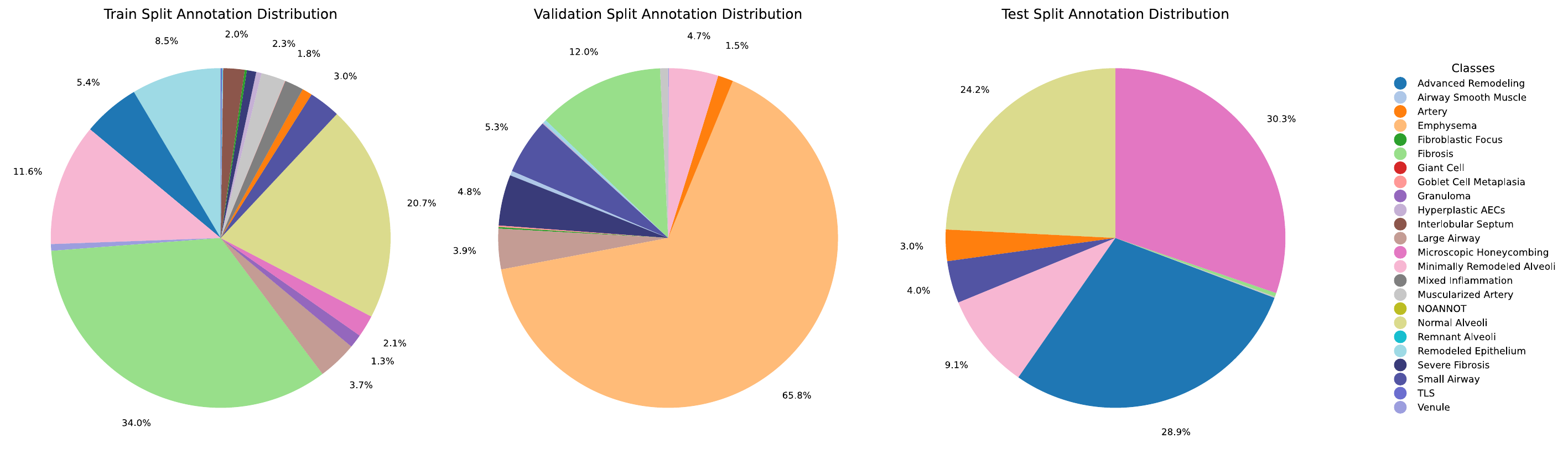}
    \caption{Label distribution in the lung cohort across train, validation, and test splits, defined based on patient stratification. The NOANNOT category is excluded from the visualization. A large distribution shift is observed between the splits.}
    \label{fig:lung_dist_shift}
\end{subfigure}

\begin{subfigure}{\textwidth}
    \includegraphics[width=\textwidth]{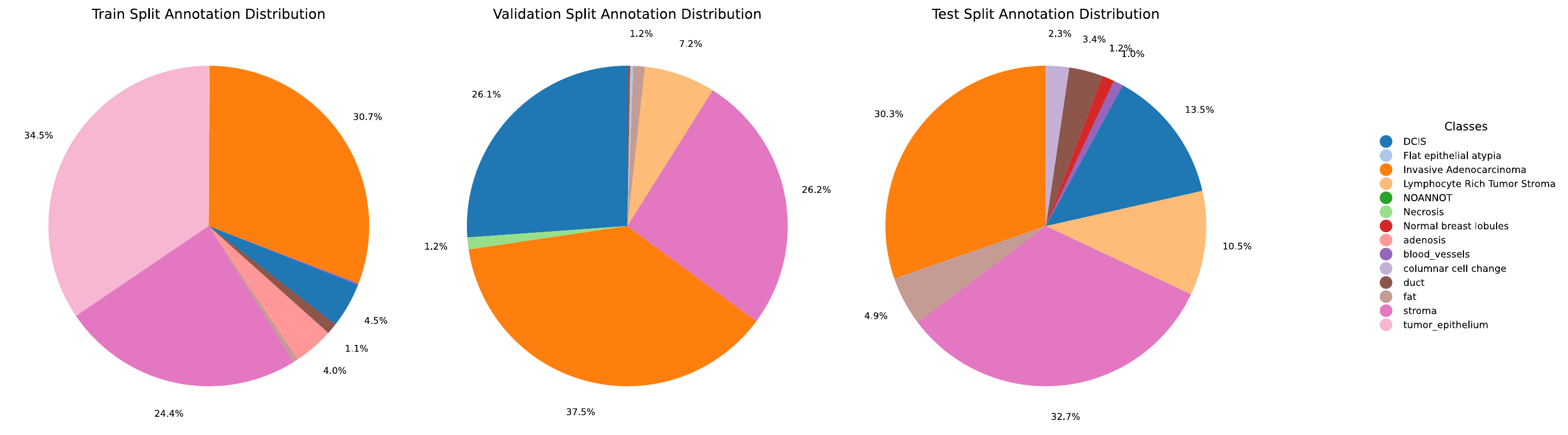}
    \caption{Label distribution in the breast cohort across train, validation, and test splits, defined based on patient stratification. The NOANNOT category is excluded from the visualization. A moderate distribution shift is observed between the splits.}
    \label{fig:breast_dist_shift}
\end{subfigure}

\begin{subfigure}{\textwidth}
    \includegraphics[width=\textwidth]{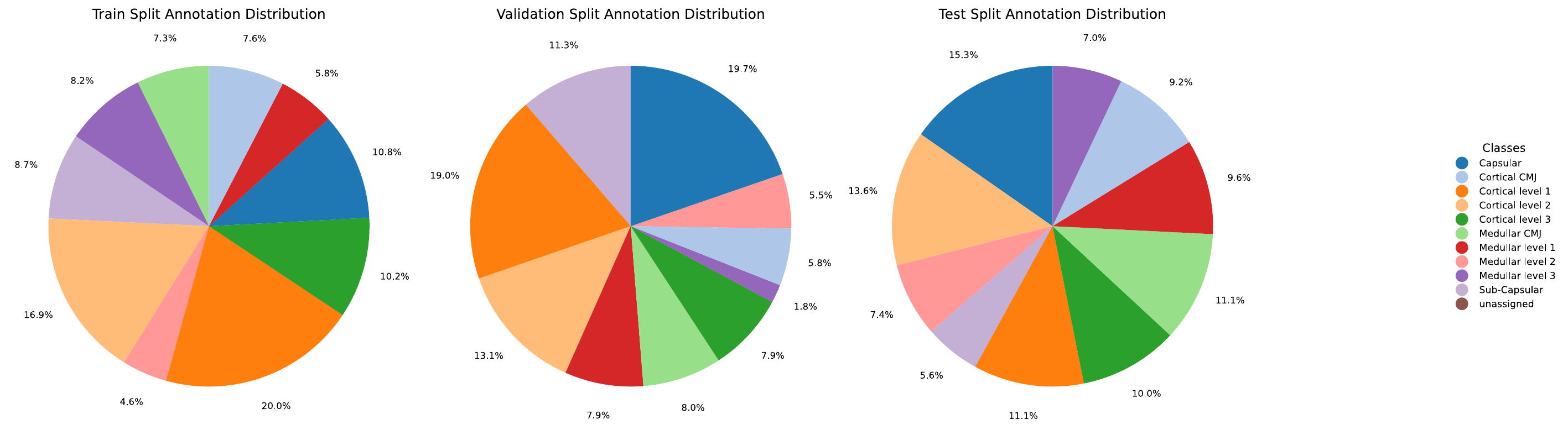}
    \caption{Label distribution in the thymus cohort across train, validation, and test splits, defined based on patient stratification. The NOANNOT category is excluded from the visualization. A small distribution shift is observed between the splits.}
    \label{fig:thymus_dist_shift}
\end{subfigure}

\caption{Label distributions in the lung, breast, and thymus cohorts across train, validation, and test splits (patient-stratified). The NOANNOT category is excluded.}
\label{fig:all_dist_shifts}
\end{figure*}

The lung dataset consists of $23$ Xenium samples from $19$ patients, covering both healthy tissue and varying severities of pulmonary fibrosis. Niche regions were annotated by expert pathologists on the H\&E-stained images, while single-cell annotations were provided at multiple levels of granularity by the original study’s authors. For our analysis, we use the broadest cell-type labels. In total, the dataset includes approximately 1 million cells across $\sim71K$ patches (Table~\ref{tab:sample-summary}). Figure \ref{fig:lung_dist_shift} shows the niche label distribution across the training, validation, and test splits, highlighting label imbalance and distribution shift between the lung samples.

The breast cancer dataset was generated and provided by 10x Genomics, the manufacturer of the Xenium technology. It includes $7$ Xenium samples from $4$ patients, capturing breast tissue affected by cancer. This dataset comprises  $\sim2.3$ million cells distributed across $\sim126K$ patches. Niche-level annotations were provided by our collaborating pathologist using the Elston-Ellis scoring system~\citep{ene}. Figure \ref{fig:breast_dist_shift} shows the niche label distribution across the training, validation, and test splits in the breast cohort.

The thymus dataset consists of $19$ Visium samples from $11$ donors, spanning fetal (post-conception weeks 11–21) and pediatric (neonate to 3 years old) tissue. This dataset was created to map T cell development in pre- and early postnatal stages. The dataset was annotated using the Cortico-Medullary Axis (CMA), a common coordinate framework developed by the dataset's original authors. As the thymus dataset is a spot-based Visium dataset, it contains transcriptomes from spots rather than individual cells. These spots are distributed across $\sim40K$ patches. Pediatric tissue samples were used for training, while fetal samples were held out for testing (Table~\ref{tab:sample-summary}). Figure \ref{fig:thymus_dist_shift} shows the niche label distribution across the training, validation, and test splits in the thymus cohort.

An overview of the dataset composition across all three cohorts is presented in Table~\ref{tab:sample-summary}, which details the number of niches, cell types, cell counts, number of patches, and split assignment per sample. Figure~\ref{fig:data_dist} further illustrates the distribution of niche types and their corresponding cell type compositions in the lung cohort, highlighting the heterogeneity within the dataset.

\begin{table*}
\small
\centering
\resizebox{\textwidth}{!}{%
\begin{tabular}{@{}lllrrrrl@{}}
\toprule
\textbf{Tissue} & \textbf{Patient} & \textbf{Sample} & \textbf{\#Niches} & \textbf{\#Cell Types} & \textbf{\#Cells} & \textbf{\#Patches} & \textbf{Split}\\
\midrule
Lung & THD0008  & THD0008    & 5  & 9  & $117,612$  & $4,316$ & Train\\
Lung & THD0011  & THD0011    & 5  & 9  & $53,544$  & $2,099$ & Val \\
Lung & TILD117  & TILD117MF  & 11 & 9  & $86,728$  & $3,831$ & Val \\
Lung & TILD117  & TILD117LF  & 8  & 9  & $65,423$  & $2,800$ & Val \\
Lung & TILD175  & TILD175    & 11 & 10 & $59,824$  & $2,489$ &Train\\
Lung & VUHD069  & VUHD069    & 4  & 9  & $36,862$  & $1,673$ & Test\\
Lung & VUHD095  & VUHD095    & 4  & 9  & $19,631$  & $808$  &Train\\
Lung & VUHD113  & VUHD113    & 2  & 9  & $20,850$  & $1,281$ &Train\\
Lung & VUHD116  & VUHD116B   & 4  & 9  & $48,213$  & $1,898$ &Train\\
Lung & VUHD116  & VUHD116A   & 3  & 9  & $19,005$  & $926$  &Train\\
Lung & VUILD102 & VUILD102LF & 5  & 10 & $45,545$  & $2,616$ &Train\\
Lung & VUILD104 & VUILD104LF & 11 & 9  & $63,792$  & $2,593$ &Train\\
Lung & VUILD105 & VUILD105MF & 6  & 9  & $39,419$  & $1,779$ &Train\\
Lung & VUILD106 & VUILD106   & 11 & 10 & $235,514$ & $9,923$ &Train\\
Lung & VUILD107 & VUILD107MF & 11 & 9  & $115,596$  & $4,392$ &Train\\
Lung & VUILD110 & VUILD110   & 5  & 9  & $117,612$  & $4,316$ &Train\\
Lung & VUILD115 & VUILD115   & 13 & 10 & $176,255$ & $7,305$ &Train\\
Lung & VUILD48  & VUILD48MF  & 8  & 9  & $52,956$  & $2,352$ &Train\\
Lung & VUILD78  & VUILD78LF  & 5  & 9  & $55,614$  & $2,227$ & Test\\
Lung & VUILD78  & VUILD78MF  & 7  & 10 & $73,135$  & $2,722$ & Test\\
Lung & VUILD91  & VUILD91LF  & 5  & 10 & $23,427$  & $1,450$ &Train\\
Lung & VUILD96  & VUILD96LF  & 8  & 9  & $76,223$  & $3,479$ &Train\\
Lung & VUILD96  & VUILD96MF  & 7  & 9  & $95,241$  & $4,034$ &Train \\
Breast & Patient 1  & Rep1   & 9  & 12 & $261,661$ & $11,122$& Val \\
Breast & Patient 1  & Rep2   & 8  & 12 & $170,749$ & $7,381$ & Val\\
Breast & Patient 2  &Sample 2& 10 & 16 & $221,766$ & $9,101$ & Test\\
Breast & Patient 3  &TENX94 & 7 & * & $545,980$ & $18,042$ & Train\\
Breast & Patient 3  &TENX96 & 7 & * & $557,757$ & $18,277$ & Train\\
Breast & Patient 4  &TENX95 & 7 & * & $917,241$ & $31,154$ & Train\\
Breast & Patient 4  &TENX97 & 7 & * & $919,345$ & $31,693$ & Train\\
Thymus & Z1 &TA11486161 & 11 & * & * & $1,445$ & Train\\
Thymus & Z2 &TA11556492 & 11 & * & * & $3,045$ & Train\\
Thymus & Z2 &TA11486164 & 11 & * & * & $2,827$ & Train\\
Thymus & Z4 &TA11486162 & 11 & * & * & $3,210$ & Train\\
Thymus & Z4 &TA11556493 & 11 & * & * & $2,963$ & Train\\
Thymus & Z4 &TA11556494 & 11 & * & * & $2,481$ & Train\\
Thymus & Z6 &TA11556496 & 11 & * & * & $2,295$ & Train\\
Thymus & Z6 &TA11556495 & 11 & * & * & $2,764$ & Train\\
Thymus & F113 &WSSS\_F\_IMMsp11604689 & 10 & * & * & $311$ & Val\\
Thymus & F114 &WSSS\_F\_IMMsp11604687 & 11 & * & * & $1,264$ & Test\\
Thymus & F114 &WSSS\_F\_IMMsp11604688 & 11 & * & * & $1,162$ & Test\\
Thymus & F130 &WSSS\_F\_IMMsp11604686 & 11 & * & * & $2,121$ & Test\\
Thymus & F130 &WSSS\_F\_IMMsp11604685 & 11 & * & * & $1,846$ & Test\\
Thymus &  F133 &WSSS\_F\_IMMsp11765867 & 11 & * & * & $1,606$ & Val\\
Thymus & F135 &WSSS\_F\_IMMsp11765870 & 11 & * & * & $2,570$ & Val \\
Thymus & U09 &WSSS\_THYst9518033 & 11 & * & * & $2,125$ & Train\\
Thymus & U11 &WSSS\_THYst9518030 & 11 & * & * & $2,384$ & Train\\
Thymus & U11 &WSSS\_THYst9142088 & 11 & * & * & $2,035$ & Train\\
Thymus & U11 &WSSS\_THYst9142089 & 11 & * & * & $2,190$ & Train\\
\bottomrule
\end{tabular}
}
\caption{Summary of samples, patients, unique niche annotations (including NOANNOT), unique cell types (including NOMAP), cell counts, number of patches, and data split (train/val/test) for the lung, breast and thymus datasets. Individual cells can be counted more than once when they intersect with multiple patches. For the thymus dataset, which is spot-based rather than cell-based, cell counts and cell-type classifications do not apply and are denoted with *. Similarly, cell-type annotations are unavailable for certain breast samples; in those cases, the number of unique cell types is also marked with *.}
\label{tab:sample-summary}
\end{table*}

\section{Impact Statement}
\label{app_impact}
This work advances multimodal learning techniques for scenarios with limited paired data. By releasing our annotated HuggingFace dataset and open-source implementation, we aim to facilitate reproducibility in multimodal research. While our framework is developed for biomedical applications, the underlying techniques could benefit other domains where data pairing is costly or impractical. Given the medical context of our work and the sensitive nature of patient-derived molecular and histological data, applications must carefully consider data privacy, informed consent, and clinical validation requirements. The integration of multiple data modalities further increases the identifiability of patient information, emphasizing the need for responsible data handling.

\end{document}